\documentclass[pdflatex,sn-mathphys-num]{sn-jnl}

\usepackage{graphicx}%
\usepackage{multirow}%
\usepackage{amsmath,amssymb,amsfonts}%
\usepackage{amsthm}%
\usepackage{mathrsfs}%
\usepackage[title]{appendix}%
\usepackage{xcolor}%
\usepackage{textcomp}%
\usepackage{manyfoot}%
\usepackage{booktabs}%
\usepackage{algorithm}%
\usepackage{algorithmicx}%
\usepackage{algpseudocode}%
\usepackage{listings}%
\usepackage{comment}
\usepackage{dcolumn}
\usepackage{mathptmx}
\usepackage{etoolbox}
\usepackage{hyperref}
\usepackage{amssymb}
\usepackage{graphicx}
\usepackage{dcolumn}
\usepackage{bm}
\usepackage{comment}
\usepackage{xcolor}
\usepackage{capt-of}
\usepackage{subcaption}

\theoremstyle{thmstyleone}%

\theoremstyle{thmstyletwo}%

\theoremstyle{thmstylethree}%
\usepackage{xcolor}

\begin{document}

\title{Jump-diffusion models of parametric volume-price distributions}

\author[1]{\fnm{Anup} \sur{Budhathoki}}

\author[2,3]{\fnm{Leonardo Rydin} \sur{Gorj\~ao}}

\author[1,4]{\fnm{Pedro G.} \sur{Lind}}

\author*[1]{\fnm{Shailendra} \sur{Bhandari}\email{shailendra.bhandari@oslomet.no}}

\affil*[1]{\orgdiv{Department of Computer Science}, \orgname{OsloMet -- Oslo Metropolitan University}, \orgaddress{\street{Pilestredet 52}, \city{Oslo}, \postcode{N-0166}, \country{Norway}}}

\affil[2]{\orgdiv{Department of Environmental Sciences, Faculty of Science}, \orgname{Open University of The Netherlands}, \orgaddress{\city{Heerlen}, \postcode{6419AT}, \country{The Netherlands}}}

\affil[3]{\orgdiv{Faculty of Science and Technology}, \orgname{Norwegian University of Life Sciences}, \orgaddress{\city{\AA s}, \postcode{1432}, \country{Norway}}}

\affil[4]{\orgdiv{School of Economics, Innovation and Technology}, \orgname{Kristiania University of Applied Sciences}, \orgaddress{\street{Kirkegata 24-26}, \city{Oslo}, \postcode{N-0153}, \country{Norway}}}


\abstract{
We present a data-driven framework to model the stochastic evolution of volume-price distribution from the New York Stock Exchange (NYSE) equities. The empirical distributions are sampled every 10 minutes over 976 trading days, and fitted to different models, namely Gamma, Inverse Gamma, Weibull, and Log-Normal distributions. Each of these models is parameterized by a shape parameter, $\phi$, and a scale parameter, $\theta$, which are detrended from their daily average behavior. The time series of the detrended parameters is analyzed using adaptive binning and regression-based extraction of the Kramers--Moyal (KM) coefficients, up to their sixth order, enabling the classification of its intrinsic dynamics. We show that  (i) $\phi$ is well described as a pure diffusion with a linear mean regression for the Gamma, Inverse Gamma, and Weibull models, while $\theta$ shows dominant jump-diffusion dynamics, with an elevated fourth- and sixth-order moment contributions;  (ii) the log-normal model shows however the opposite: $\theta$ is predominantly diffusive, with $\phi$ showing weak jump signatures; (iii) global moment inversion yields jump rates and amplitudes that account for a large share of total variance for $\theta$, confirming that rare discontinuities dominate volatility.
}

\keywords{Jump-diffusion modeling, Kramers–Moyal coefficients, Volume-price dynamics, Financial market volatility}

\maketitle

\section{Introduction and motivation}
\label{sec:intro}

Financial markets exhibit complex dynamics characterized by continuous fluctuations and abrupt, discontinuous shifts in asset prices. Traditional stochastic models, such as Geometric Brownian motion or Langevin dynamics~\cite{sheldon2014Introduction}, effectively capture smooth price evolution but struggle to account for extreme events like market crashes or liquidity shocks~\cite{black1973pricing, rocha2016evolution}. These events, often manifesting as heavy-tailed distributions in empirical data, necessitate models that incorporate both continuous and discrete components~\cite{persio2016jump}. Jump-diffusion models address this limitation by combining a continuous diffusion process with a Poisson-driven jump component, enabling the representation of both regular market movements and sudden price shifts~\cite{merton1976jump}.
Described by the stochastic differential equation
\begin{equation}
\mathrm{d}X_t = a(X_t) \mathrm{d}t + b(X_t) \mathrm{d}W_t + \xi dJ_t,
\label{eq:jd}
\end{equation}
where $a(X_t)$ is the drift, $b(X_t)$ the diffusion coefficient, $W_t$ a Wiener process, $\xi$ the jump size, and $J_t$ a Poisson process with intensity $\lambda$, these models capture the heavy-tailed and nonstationary nature of financial time series~\cite{RydinGorjao2023}. In parallel, superstatistics~\cite{Beck2003, Beck2004}, which models non-equilibrium systems as a superposition of different local equilibrium dynamics, provides an alternative or complementary approach. By accounting for slowly varying parameters, such as volatility or temperature-like variables, superstatistical frameworks are well-suited to capture the intermittent and multiscale nature of financial market fluctuations~\cite{klages2008anomalous, estevens2017stochastic}.

L\'evy processes provide a general framework for modeling continuous and jump dynamics via independent, stationary increments~\cite{Applebaum_2009}. Parametric L\'evy models (variance-gamma~\cite{madan1990variance}, Carr-Geman-Madan-Yor (CGMY) model~\cite{carr2002fine}, Normal Inverse Gaussian (NIG)~\cite{https://doi.org/10.1111/1467-9469.00045}) pre-specify the jump measure $\nu(dz)$ and calibrate parameters from data; the Merton model of Eq.~\eqref{eq:jd} is a special case with Poisson jumps. In contrast, the Kramers-Moyal approach is nonparametric, reconstructing drift, diffusion, and jump characteristics directly from conditional moments without assuming a specific L\'evy measure~\cite{RydinGorjao2023, tabar2019analysis}, which is advantageous when jump dynamics are unknown.
\begin{figure*}[t]
    \centering
    \includegraphics[width=\linewidth]{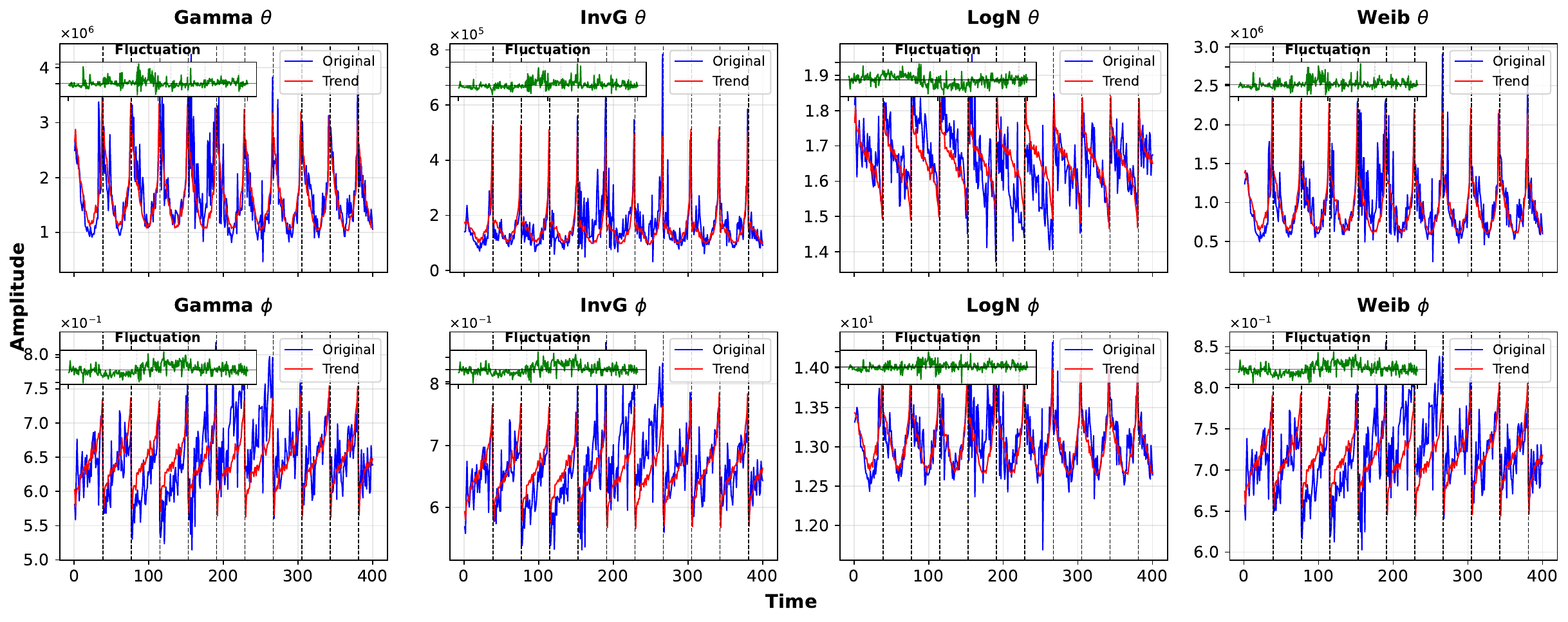}
    \caption{Daily variation and trends in $\theta$ and $\phi$ for the Gamma, Inverse Gamma, Weibull, and Log-Normal distributions. Each subplot shows original values (blue), 21-day moving average (red), and fluctuations (green, inset). Vertical lines mark daily cycles every 38 points.}
    \label{fig:Gamma_LeastSquare_theta}
\end{figure*}

In practical applications, Kramers-Moyal (KM) estimators are effective for identifying stochastic processes in noisy datasets, particularly for detecting discontinuous jumps in a system driven by stochastic forcing, as shown in paleoclimate ice core data showing heavy-tailed dynamics similar to financial markets~\cite{riechers2023discontinuous}. However, their application to real-world financial data requires careful consideration of assumptions such as the Markov property, time-homogeneity, dense state space sampling, and sampling intervals short relative to system dynamics, as well as challenges like finite sampling effects, nonstationarity, and the presence of outliers, which can distort coefficient estimates~\cite{riechers2023discontinuous, friedrich2011approaching}. Complementary methods for handling nonstationary red noise, such as filtering to autocorrelation structures or power spectral densities, further enhance robustness in real-world noisy environments~\cite{PhysRevX.14.021037}. Robust preprocessing and sufficient data resolution are critical to ensure accurate detection of jump contributions versus continuous dynamics in intraday trading environments, as seen in applications to turbulent flows and other complex systems~\cite{friedrich2011approaching, RENNER_PEINKE_FRIEDRICH_2001}. The successful application of KM methods to financial data requires careful attention to several critical assumptions and practical considerations, including the Markov property, time-homogeneity, finite-lag effects, and the presence of outliers, which are systematically addressed through our four-step analysis pipeline in Section 2.
This work aims to quantify the role of jumps in intraday volume-price fluctuations using high-frequency financial data. Our objectives are: a) Estimate KM coefficients to assess the necessity of jump-diffusion models over pure diffusion frameworks. b) Evaluate the contribution of jumps to market dynamics and their implications for risk management.
Integrating stochastic process theory with computational techniques, we provide a robust framework for modeling financial volatility, advancing both theoretical understanding and practical risk assessment.

\section{Data and methods}
\label{sec:data_methods}
We analyze the volume-price of 1\,750 NYSE-listed companies, with a sampling frequency of 0.1 min$^1$. After removing all the after-hours trading and discarding all the days with recorded errors, our dataset contains 17708 data points for each company covering a total period of 976 days~\cite{rocha2016evolution, estevens2017stochastic}. The filtering process is detailed as follows. Each raw file contains 136,002 rows 
recorded at 10-minute intervals over all 24 hours. However, only the 6.5 trading hours (09:10--15:30) yield valid observations; remaining entries are zero or missing. Extracting the non-zero trading blocks gives $466 \text{ trading days} \times 38 \text{ intraday bars} = 17\,708 \text{ observations}$ per series. We then apply a $10\sigma$ outlier rule: for each series, any trading day where at least one value exceeds the series mean plus ten standard deviations is flagged. The per-series results are shown in Table~\ref{tab:outlier_days}. The union gives 4 outlier days $\{123, 353, 375, 424\}$, which are removed from every series to ensure identical trading-day calendars, leaving 17,556 observations (462 days) per series. Finally, a 21-day centered moving average (excluding the current day) is subtracted and the first and last 10 days are clipped to avoid edge effects, yielding the final detrended series of $N = 16\,796$ observations (442 days) 
per series.

\begin{table}[t]
\centering

\caption{Per-series outlier detection using $10\sigma$ rule. Threshold = 
$\mu + 10\sigma$ for each series.}
\label{tab:outlier_days}
\small
\begin{tabular}{lrrl}
\hline
\textbf{Series} & \textbf{Mean} & \textbf{Threshold} ($\mu+10\sigma$) & 
\textbf{$N^{\text{th}}$ outlier day
} \\
\hline
Gamma-$\phi$   & 0.648         & 1.20              & 353 \\
Gamma-$\theta$ & $1.59 \times 10^6$ & $8.21 \times 10^6$ & 123 \\
InvG-$\phi$    & 0.671         & 1.28              & 353, 375 \\
InvG-$\theta$  & $1.57 \times 10^5$ & $1.09 \times 10^6$ & 424 \\
LogN-$\phi$    & 13.0          & 17.1              & (none) \\
LogN-$\theta$  & 1.65          & 2.69              & 123 \\
Weib-$\phi$    & 0.720         & 1.18              & 353 \\
Weib-$\theta$  & $9.14 \times 10^5$ & $4.85 \times 10^6$ & (none) \\
\hline
\end{tabular}
\end{table}

Each observation contains a price $p(t)$ and a volume $v(t)$, and we study the volume-price variable $s(t)=p(t)v(t)$ as a composite measure of monetary flow and trading intensity. We fit four standard positive-support distribution families to the volume-price $s(t)$ at the intraday level: Gamma ($f_G$), Inverse Gamma ($f_{IG}$), Weibull ($f_W$), and Log-Normal ($f_{LN}$). Each family is parameterized by a shape $\phi$ and a scale $\theta$ (for $f_{LN}$, $\phi$ and $\theta$ denote the log-location and log-scale). The probability density functions are:

\begin{equation}
\begin{aligned}
{f_G:}\quad
&f(x;\phi,\theta)=\frac{1}{\Gamma(\phi)\,\theta^\phi}\,x^{\phi-1}\exp(-x/\theta) \\
{f_{IG}:}\quad
&f(x;\phi,\theta)=\frac{\theta^\phi}{\Gamma(\phi)}\,x^{-\phi-1}\exp(-\theta/x) \\
{f_W:}\quad
&f(x;\phi,\theta)=\frac{\phi}{\theta^\phi}\,x^{\phi-1}\exp\!\left(-(x/\theta)^\phi\right) \\
{f_{LN}:}\quad
&f(x;\phi,\theta)=\frac{1}{x\,\theta\sqrt{2\pi}}
\exp\!\left(-\frac{(\ln x-\phi)^2}{2\theta^2}\right)
\end{aligned}
\end{equation}

All densities are defined for $x>0$. For $f_G$, $f_{IG}$, and $f_W$, $\phi,\theta>0$.
For $f_{LN}$, $\phi=\mu\in\mathbb{R}$ and $\theta=\sigma>0$ such that $\ln x\sim\mathcal{N}(\mu,\sigma^2)$.
The parameters $(\phi(t),\theta(t))$ are estimated by maximum likelihood for each intraday time index and day. To focus on stochastic fluctuations, we detrend each parameter time series with a 21-day moving average, producing $\phi'(t)$ and $\theta'(t)$. Sensitivity to the detrending window for 10 and 42 days is reported in Appendix~E (Table~\ref{tab:window_sensitivity_ratio}).

\begin{figure}
    \centering
    \begin{tabular}{@{}c@{\vspace{0.01\linewidth}}c@{}}
    \fbox{\includegraphics[width=0.475\linewidth]{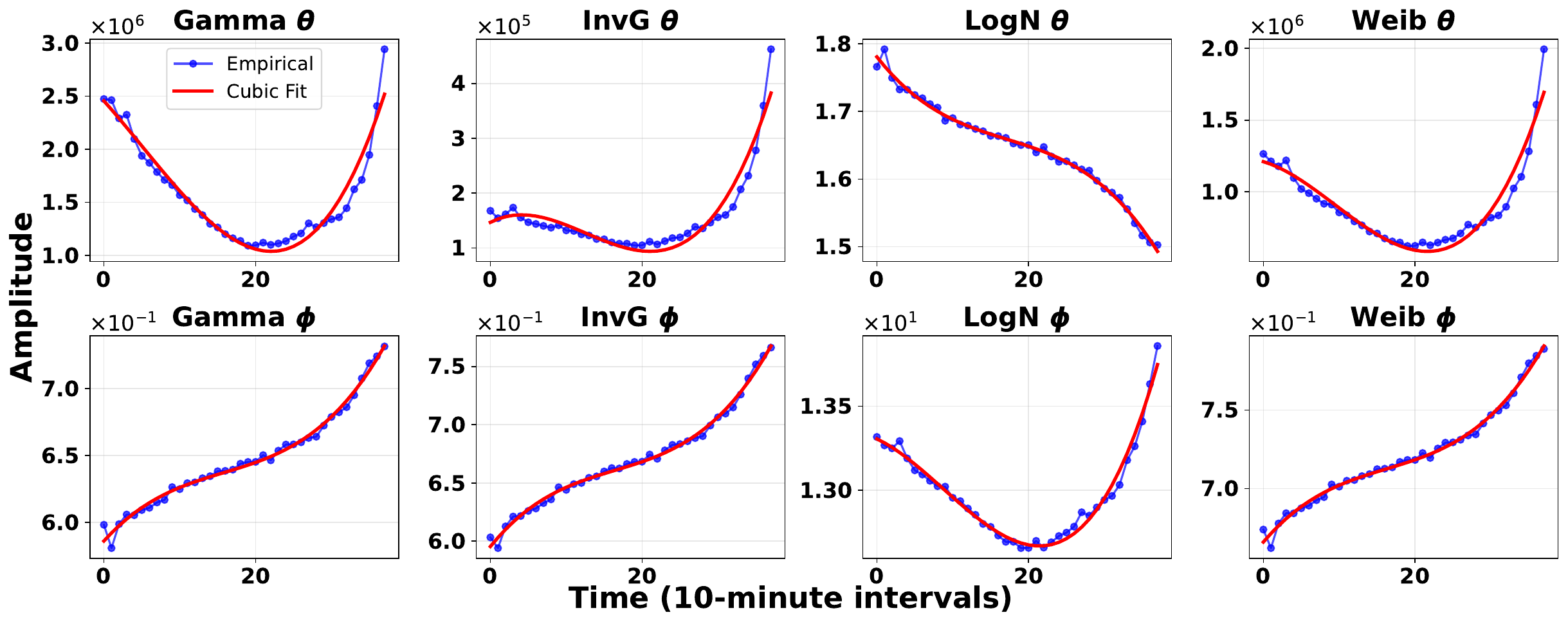}}
    \fbox{\includegraphics[width=0.475\linewidth]{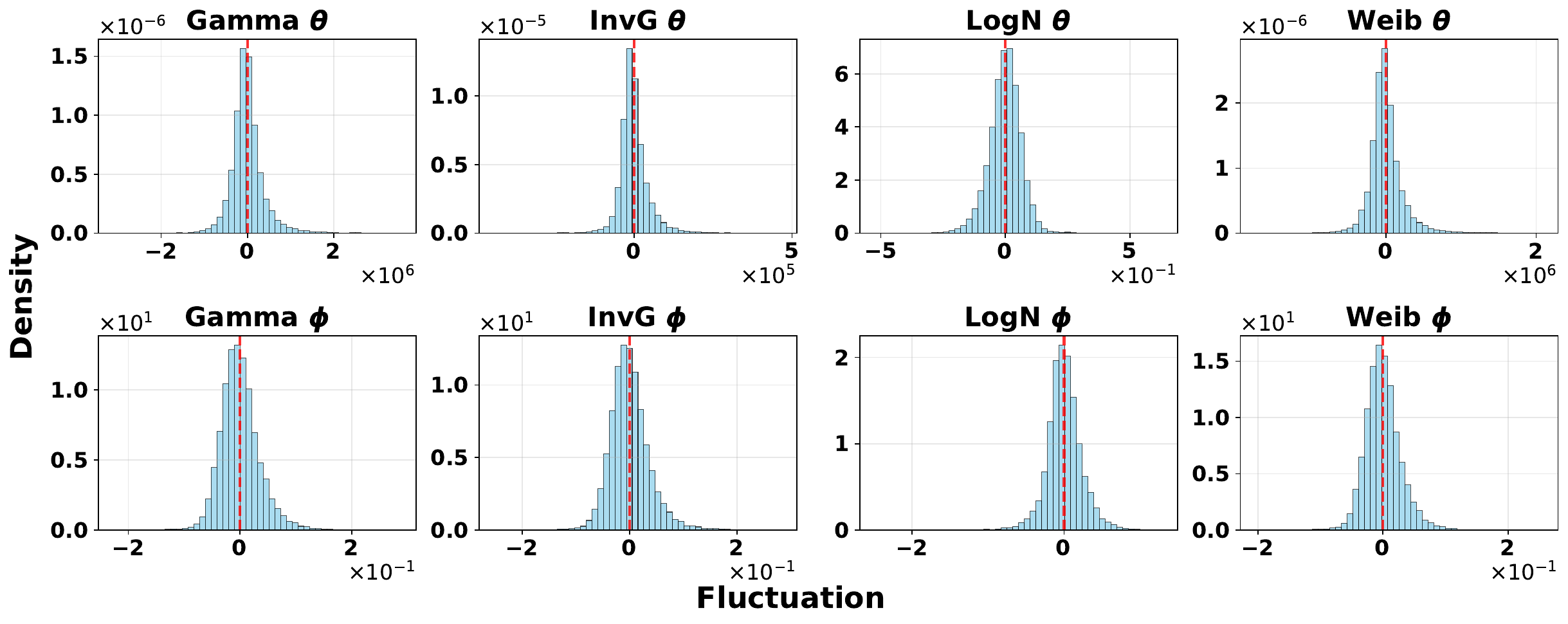}}    
    \end{tabular}
    \caption{Left: average intraday profiles of $\theta$ (first row) and $\phi$ (second row) for $f_G$, $f_{IG}$, $f_{LN}$ and $f_W$, computed over 976 trading days at 10-minute resolution. Markers show sample means at each intraday time index; red curves are cubic polynomial fits summarizing the deterministic daily pattern.  Right: empirical distributions of the detrended fluctuations $\theta'(t)$ and $\phi'(t)$ for the same four families.}
    \label{fig:cubic_fitting}
\end{figure}

Weak stationarity of the detrended series $\phi'(t)$ and $\theta'(t)$ is assessed with Augmented Dickey--Fuller (ADF) and Kwiatkowski--Phillips--Schmidt--Shin (KPSS) tests~\cite{stats6020040, 10.1063/1.4914547}. The ADF test (null: unit root) rejects nonstationarity with $p<10^{-4}$, and the KPSS test (null: stationarity) yields $p=0.1$ across all eight series ( $f_G$, $f_{IG}$, $f_{LN}$, and $f_W$ for both $\phi$ and $\theta$), supporting weak stationarity~\cite{dickey1979distribution, kwiatkowski1992testing}.
Detailed results are reported in \emph{Table~I of the Supplementary Information}. Fig.~\ref{fig:Gamma_LeastSquare_theta} summarizes the raw parameter trajectories and their intraday structure for all four distribution families. Each panel shows the original series (blue), a 21-day moving average (red) that captures slow seasonal drift, and the corresponding fluctuation series (green, inset) after removing the moving average. Vertical guidelines mark the 38 intraday sampling points. The scale parameter \(\theta(t)\) displays pronounced daily cycles with large amplitude and occasional bursts, indicating strong time-of-day effects and episodic variability. The shape parameter \(\phi(t)\) varies more smoothly with a smaller amplitude and weaker daily modulation. This contrast is consistent across $f_G$, $f_{IG}$, $f_{LN}$, and $f_W$, and it motivates detrending before stochastic analysis. Fig.~\ref{fig:cubic_fitting} characterizes the deterministic intraday pattern and the statistics of the detrended residuals. In the top two rows, we average \(\theta(t)\) and \(\phi(t)\) over trading days at each of the 38 intraday points and fit a cubic polynomial to the mean profile. The cubic polynomial captures a typical U-shaped day for \(\theta\) with a strong open, midday lull, and late-day rise, while \(\phi\) shows a flatter, low-amplitude profile.

Subtracting these fits yields fluctuation series \(\theta'(t)\) and \(\phi'(t)\) that are free of visible intraday structure.
The bottom two rows show histograms of these fluctuations.
The \(\theta'\) distributions are wider with heavier tails, consistent with jump-prone dynamics, whereas the \(\phi'\) distributions are narrower and closer to symmetric, consistent with predominantly diffusive behavior.
These features align with the KM diagnostics reported later.

To characterize continuous and discontinuous dynamics, we employ a jump-diffusion model~\cite{merton1976jump} as
\begin{equation}\label{eq:jd1}
\mathrm{d}X_t = a(X_t) \mathrm{d}t + b(X_t) \mathrm{d}W_t + \xi \mathrm{d}J_t,
\end{equation}
where $a(X_t)$ is the drift, $b(X_t)$ the diffusion coefficient, $W_t$ a Wiener process, $\xi$ the jump size, and $J_t$ a Poisson process with intensity $\lambda$.
For a sampling interval $\Delta t$, define the conditional moments of increments over lag $\tau$ (in time steps) as
\begin{equation}
k^{(m)}(x,\tau)=\mathbb{E}\left[(X(t+\tau)-X(t))^m \mid X(t)=x\right].
\end{equation}
The infinitesimal moments $M^{(m)}(x)$ are obtained from the linear slope of $k^{(m)}(x,\tau)$ versus $\tau\Delta t$ as $\tau\to 0$~\cite{tabar2019analysis}.
For jump-diffusion processes, the jump variance $\sigma_\xi^2(x)$, intensity $\lambda(x)$, and diffusion $b^2(x)$ can be estimated from corrected moments~\cite{anvari2016disentangling} as
\begin{equation}
\begin{aligned}\label{eq:jump_params_corrected}
\sigma_\xi^2(x) = \frac{M^{(6)}(x)}{5\,M^{(4)}(x)}, \;\; \lambda(x) = \frac{M^{(4)}(x)}{3\,\sigma_\xi^4(x)}, \;\; b^2(x) &= M^{(2)}(x) - \lambda(x)\,\sigma_\xi^2(x). 
\end{aligned}
\end{equation}
For a full derivation, see Ref.~\cite{tabar2019analysis}. 
Stochastic analysis of $\phi'(t)$ and $\theta'(t)$ follows a four-step pipeline to estimate KM coefficients~\cite{RydinGorjao2023, tabar2019analysis}: (1) Markov property verification, (2) adaptive state-space discretization, (3) computation of raw conditional moments, and (4) extraction of infinitesimal moments with finite-lag corrections.

\textbf{Markov property verification:}
We test the Markov property, $p(x_{t+\tau} \mid x_t, x_{t-1}, \ldots) = p(x_{t+\tau} \mid x_t)$ \cite{gardiner2009stochastic}, using the conditional entropy $H(x_{t+\tau}\mid x_t)$ and the partial autocorrelation function (PACF) \cite{box2015time}.
The conditional entropy is computed from joint histograms of $(x_t,x_{t+\tau})$ pairs as
with $p(x_{t+\tau} \mid x_t)=p(x_t,x_{t+\tau})/p(x_t)$. 
The Markov time $\tau_M$ is the smallest $\tau$ for which the entropy slope $\Delta H(\tau)=H(\tau+1)-H(\tau)$ falls below $0.005$.
As a complementary check, the PACF is estimated up to lag 20 via the Yule--Walker method; $\tau_M$ is the first lag whose PACF lies within $\pm 1.96/\sqrt{N}$, with $N\approx 1.7{\times}10^{4}$ effective samples~\cite{box2015time, gong1992markov}.

\textbf{Adaptive state space discretization:}
We discretized the state variable into bins optimized for conditional moment stability.
Classical histogram rules (Freedman--Diaconis, Scott, Sturges, and Doane) either over-smooth cores or sparsify tails~\cite{RydinGorjao2023}.
We implement zone-adaptive binning, partitioning the state space into core ($|x| < \sigma$), shoulder ($\sigma \leq |x| < 2\sigma$), and tail ($|x| \geq 2\sigma$) regions, where $\sigma$ is the sample standard deviation.
Bins are sized to ensure 350--400 points in core, 250--300 in shoulder, and 150--200 in tail regions, eliminating empty bins while stabilizing higher-order moment estimates.

\textbf{Raw conditional moments:}
For each state space bin $b$ centered at $x_b$ and time lag $\tau$ (in time steps), raw conditional moments are computed as
 $K^{(n)}(x_b, \tau) = \frac{1}{N_b} \sum_{i \in b} [\Delta x_i(\tau)]^n, \quad n = 1, 2, \ldots, 6,$
where $N_b$ is the number of data points in bin $b$, $\Delta x_i(\tau) = x(t_i + \tau) - x(t_i)$ is the increment over lag $\tau$, and moments are computed for $\tau \in \{1, 2, 3, 4, 5, 6\}$ time steps. For all datasets analyzed, the Markov time $\tau_M$ ranges between 5 and 9 time steps, 
as determined from conditional entropy analysis. We therefore consistently use 
$\tau \in \{1, 2, 3, 4, 5, 6\}$ for regression across all datasets and distribution 
families. This range ensures $\tau < \tau_M$ (avoiding spurious correlations from memory effects) while providing sufficient $\tau$ points for regression, as required by Taylor series expansion convergence~\cite{RydinGorjao2023}.

\textbf{Infinitesimal moments via regression and corrections:}
The raw infinitesimal moments $M^{(n)}(x)$ obtained from the linear regression 
are mixed moments: each $M^{(n)}$ contains contributions from all lower orders. 
For example, $M^{(2)}$ includes both the true diffusion term and a spurious 
contribution from the squared drift $(M^{(1)})^2$. Similarly, $M^{(4)}$ mixes the 
true fourth-order jump signature with products of lower-order moments. The 
correction formulas remove these cross-contaminations by applying classical 
cumulant-to-moment relations expressed via Bell polynomials. Infinitesimal conditional moments $M^{(n)}(x)$ are extracted by regressing normalized raw moments against time lag.
For each bin and order $n$, we perform ordinary least squares regression:
\begin{equation}\label{eq:regression}
\frac{K^{(n)}(x, \tau)}{\tau \Delta t} = M^{(n)}(x) + \beta (\tau \Delta t) + \epsilon,
\end{equation}
where $\Delta t$ is the sampling interval, the intercept $M^{(n)}(x)$ represents the infinitesimal moment in the limit $\tau \to 0$, and the slope $\beta$ captures systematic finite-$\tau$ effects~\cite{tabar2019analysis}. To correct for finite-lag biases, we apply the KM correction formulas~\cite{RydinGorjao2021arbitrary,RydinGorjao2023}, transforming $M^{(n)}$ to corrected infinitesimal moments $F_n$ as
\begin{equation}\label{eq:corrections}
\begin{aligned}
F_1 &= M_1,\; F_2 = M_2 - M_1^2, \; F_3 = M_3 - 3M_1M_2 + 3M_1^3, \\
F_4 &= M_4 - 4M_1M_3 + 18M_1^2M_2 - 3M_2^2 - 15M_1^4, \\
F_5 &= M_5 - 5M_1M_4 + 30M_1^2M_3 - 150M_1^3M_2 + 45M_1M_2^2 - 10M_2M_3 + 105M_1^5, \\
F_6 &= M_6 - 6M_1M_5 + 45M_1^2M_4 - 300M_1^3M_3 + 1575M_1^4M_2 - 675M_1^2M_2^2 + 180M_1M_2M_3 \\
&\quad + 45M_2^3 - 15M_2M_4 - 10M_3^2 - 945M_1^6.
\end{aligned}
\end{equation}

The corrected quantities $F_n$ isolate the genuine $n$-th order contribution, so 
that $D^{(n)}(x) = F_n(x)/n!$ faithfully represents the drift ($n=1$), diffusion 
($n=2$), and jump ($n=4,6$) coefficients of the underlying stochastic process. 
Without these corrections, especially for higher orders $n \geq 4$, the moments 
become severely distorted, leading to spurious jump signatures or missed jump 
activity. This correction step is fundamental to obtaining reliable KM coefficients 
from finite datasets. Then the drift $D^{(1)}(x)$ and diffusion $D^{(2)}(x)$ coefficients govern continuous dynamics, while higher-order coefficients $D^{(4)}(x)$ and $D^{(6)}(x)$ quantify jump contributions. Following Pawula's theorem~\cite{Risken1996}, if $D^{(4)}(x)$ is negligible relative to $D^{(2)}(x)$ across all state bins (specifically, if the diagnostic ratio $D^{(4)}/D^{(2)} < 0.1$), the process is purely diffusive and adequately described by a Langevin equation.
Otherwise, jump-diffusion dynamics are indicated.


\section{Results}
\label{sec:results}

We present the empirical findings from the analysis of the detrended $\phi'(t)$ and $\theta'(t)$ time series for $f_G$, $f_{IG}$, $f_{LN}$, and $f_W$ volume-price distributions, using the stochastic modeling pipeline described in Section~\ref{sec:data_methods}.
To characterize the temporal structure of the detrended fluctuation time series $\phi'(t)$ and $\theta'(t)$ for datasets ($f_G$, $f_{IG}$, $f_{LN}$, and $f_W$ distributions for $\phi$ and $\theta$), we analyzed three diagnostics:
ACF, which measures linear correlation between values at different lags;
PACF, which isolates correlations at each lag after removing shorter-lag effects;
and the conditional entropy $H(x_{t+\tau} \mid x_t)$, which quantifies uncertainty in the future state given the present~\cite{gong1992markov, box2015time}.
These analyses confirm the approximate first-order Markov property and identify the minimum Markov lag $\tau_\mathrm{M}$ beyond which memory effects are negligible, enabling reliable conditional moment estimation in subsequent analyses.

For all datasets, ACF and PACF decay rapidly, typically falling below 95\% confidence bounds ($\pm 1.96/\sqrt{N}$, $N \approx 17{,}000$) within 7--18 lags, while conditional entropy stabilizes (slope $\Delta H < 0.005$) at lags of 5--9, indicating Markov times $\tau_\mathrm{M}$ between 5 and 9 time steps. The results validate the use of short, lags $\tau \in \{1, 2, 3, 4, 5, 6\}$ lags for computing conditional moments, ensuring the process is memoryless as required for KM analysis~\cite{RydinGorjao2023, tabar2019analysis}.

To distinguish between diffusive and jump-diffusion dynamics, we estimate raw conditional moments $K^{(n)}(x, \tau)$ for orders $n = 1, 2, 4, 6$ across the state space using adaptive binning.
If the fourth-order raw conditional moments have hills and valleys and their magnitude is comparable to the second-order conditional moments, then we can tell that it suggests a jump diffusion process.  

\begin{figure*}
    \centering
    \begin{tabular}{@{}c@{\hspace{0.01\linewidth}}c@{}}
    \fbox{\includegraphics[width=0.475\linewidth]{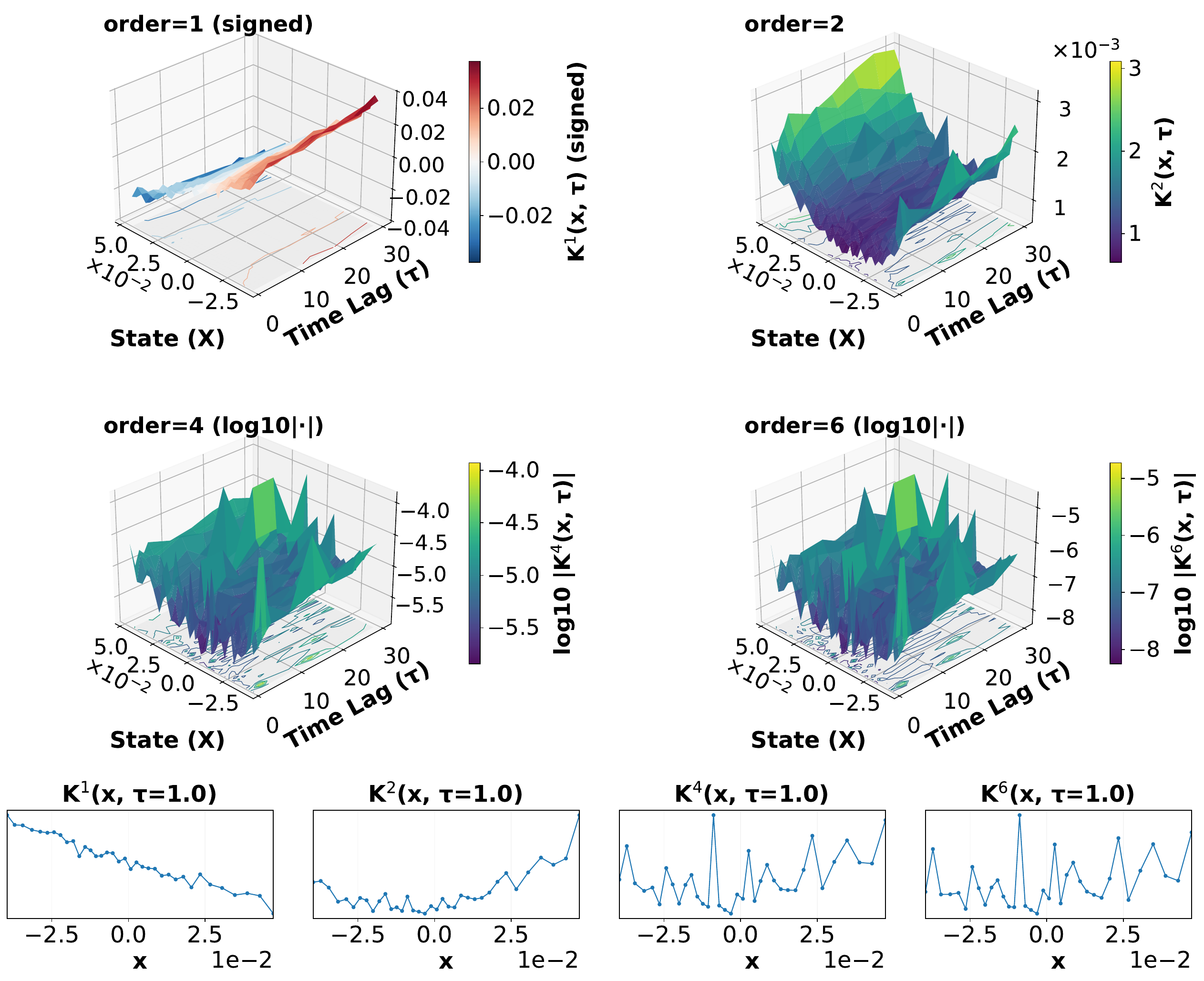}} &
    \fbox{\includegraphics[width=0.475\linewidth]{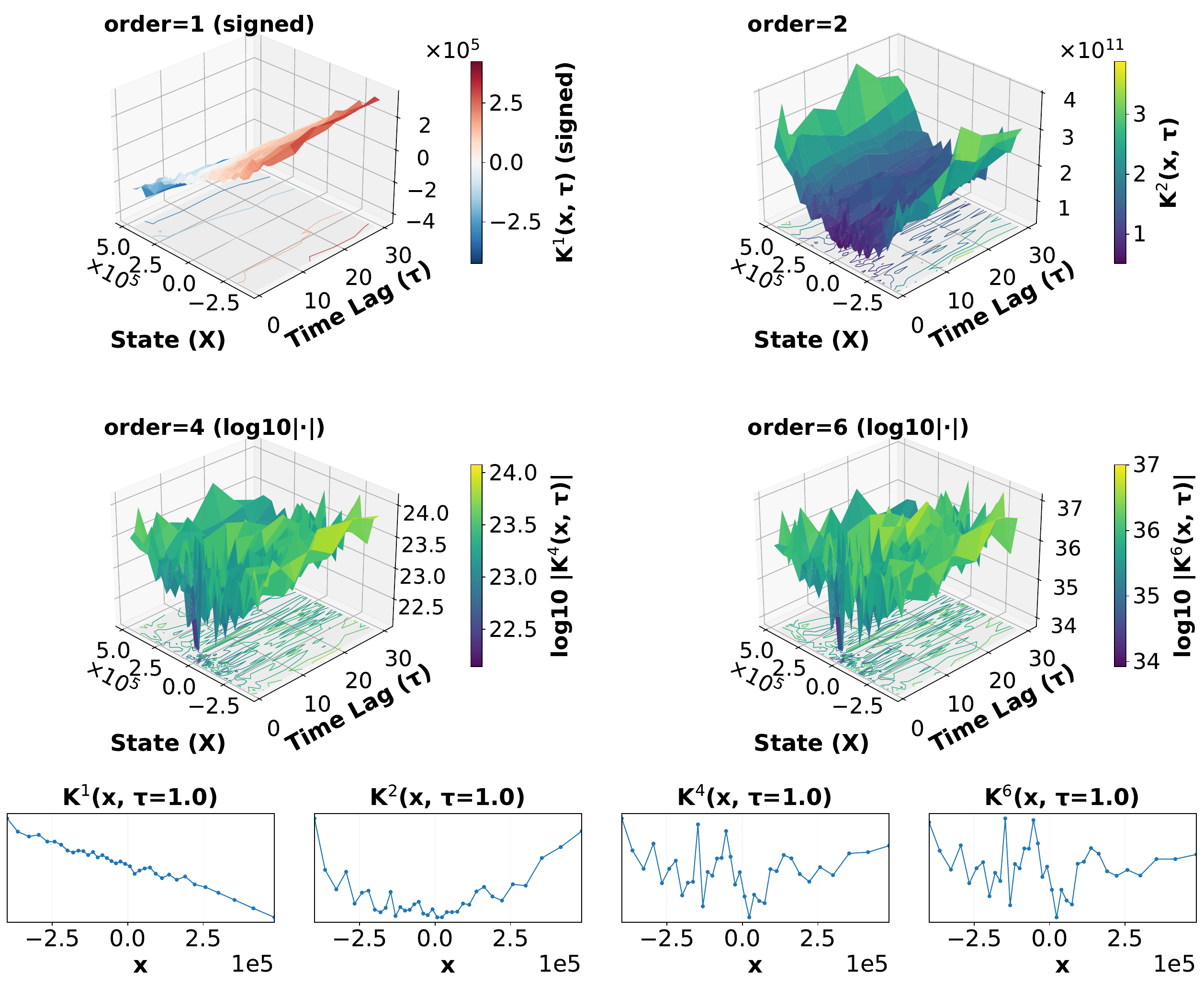}}
    \end{tabular}
    \caption{Raw conditional moments $K^{(n)}(x,\tau)$ for $f_G(\phi)$ (left) and $f_G(\theta)$ (right). Rows 1--2: 3D surfaces with contour projections for $n=1$ (signed) and $n=2$ (positive). Rows 3--4: $\log_{10}|K^{(4)}|$ and $\log_{10}|K^{(6)}|$. $f_G(\phi)$ shows flat higher-order structure consistent with diffusion; $f_G(\theta)$ shows state-localized peaks consistent with jump-diffusion. \emph{Analogous figures for $f_{IG}$, $f_{LN}$, and $f_W$ appear in Appendix~\ref{app:conditional_moments}.}}
    \label{fig:gamma_phi_theta_moments_3d}
\end{figure*}

Fig.~\ref{fig:gamma_phi_theta_moments_3d} shows raw conditional moments $K^{(n)}(x,\tau)$ for $f_G(\phi)$ (left) and $f_G(\theta)$ (right), over state $x$ and lag $\tau$.
Orders $n=1,2$ are plotted in linear scale (signed for $n=1$), while $n=4,6$ use $\log_{10}|K^{(n)}|$ to compress the dynamic range.
For $n=1$, the surface is nearly $\tau$-invariant and changes sign across $x=0$, indicating mean-reverting drift.
For $n=2$, the positive surface varies smoothly in $x$, reflecting diffusive strength.
Higher orders separate the two datasets: $f_G(\phi)$ shows flat, low-amplitude $K^{(4)}$ and $K^{(6)}$ with no localized structure (diffusion), whereas $f_G(\theta)$ exhibits state-localized peaks that persist for small $\tau$ (jump activity).
These visual cues agree with the dimensionless diagnostics (not shown): small, weakly $x$-dependent conditional excess kurtosis and $D^{(4)}/D^{(2)}<0.1$ for $f_G(\phi)$, versus large, $x$-dependent values for $f_G(\theta)$.
The corresponding conditional-moment figures for $f_{IG}$, $f_{LN}$ and $f_W$ are provided in Appendix~\ref{app:conditional_moments}.
In addition, the conditional moment analysis for $f_{IG}$, $f_{LN}$, and $f_W$ follows the same methodology and is summarized in the Appendix Table~\ref{tab:combined_summary}. 

\begin{figure}[t]
    \centering
    \begin{tabular}{@{}c@{\hspace{0.01\linewidth}}c@{}}
    \fbox{\includegraphics[width=0.47\linewidth]{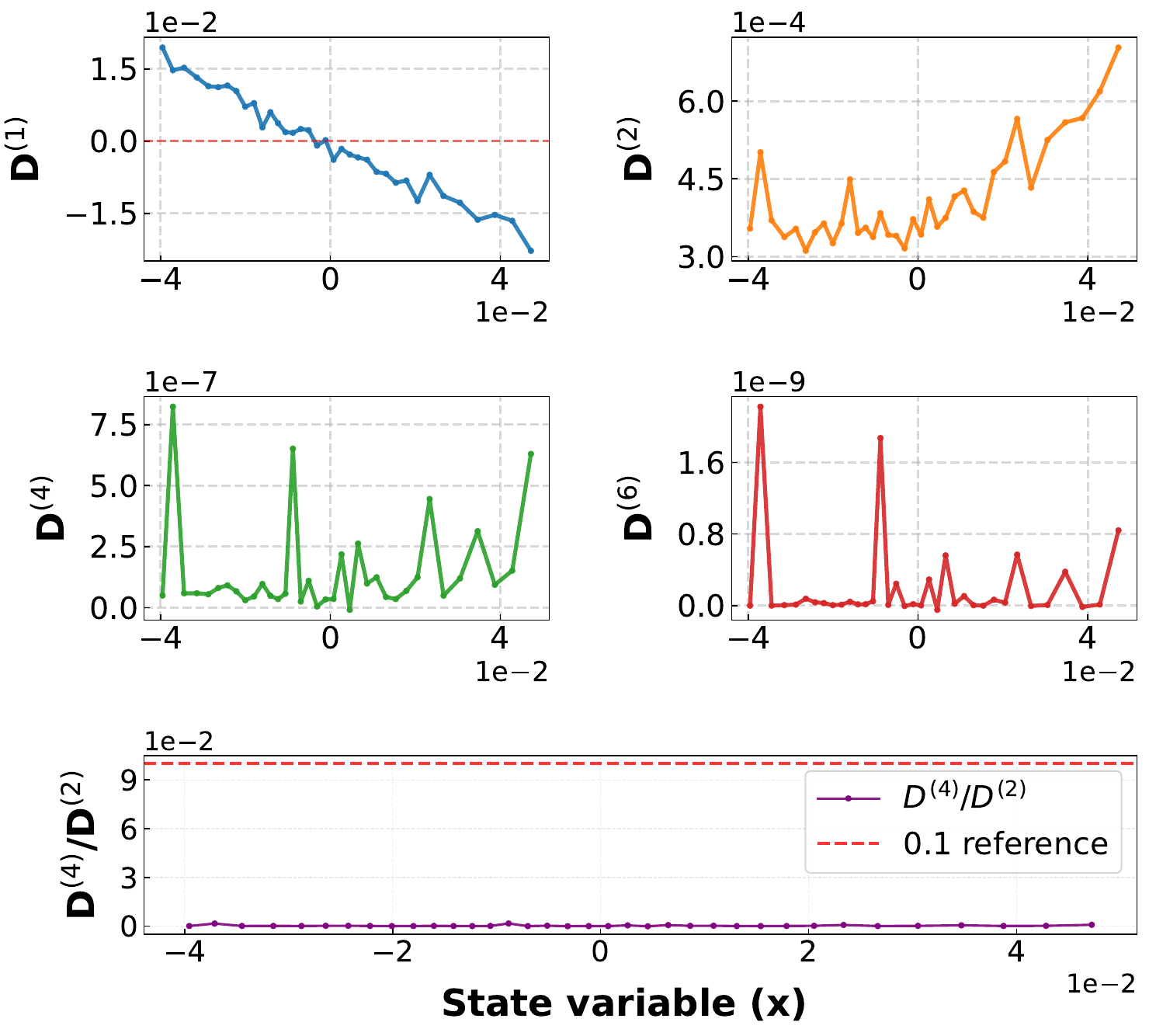}} 
    \fbox{\includegraphics[width=0.47\linewidth]{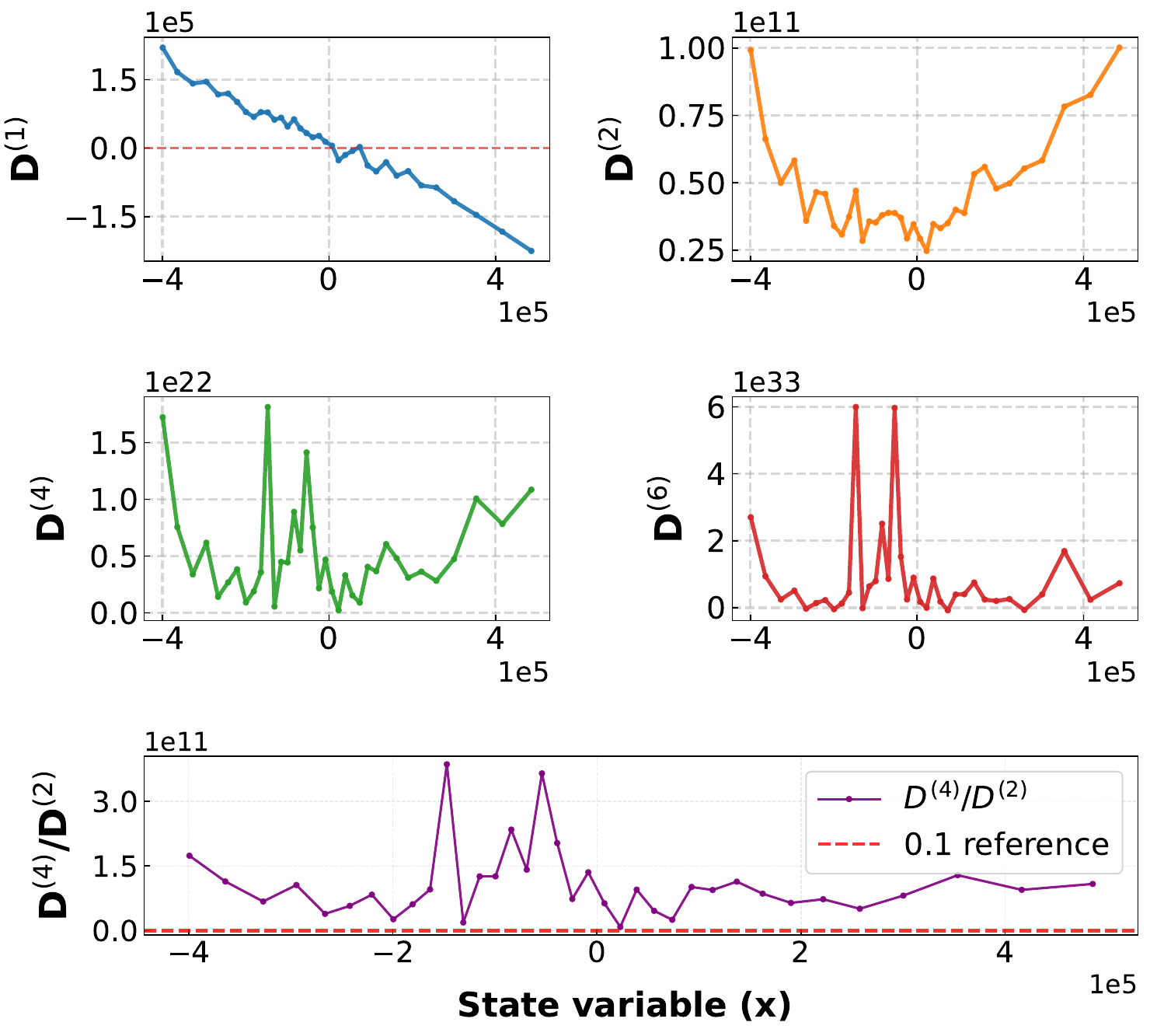}}
    \end{tabular}
    \caption{KM coefficients \(D^{(n)}(x)\) for $f_G(\phi)$ (left) and $f_G(\theta)$ (right). Each panel shows \(D^{(1)}(x)\), \(D^{(2)}(x)\), \(D^{(4)}(x)\), \(D^{(6)}(x)\), and the diagnostic ratio \(R(x)=D^{(4)}(x)/D^{(2)}(x)\) with a reference line at \(0.1\). $f_G(\phi)$ exhibits linear mean-reverting drift, stable diffusion (\(\sim 10^{-4}\)), and negligible higher-order terms (\(D^{(4)}\sim 10^{-7}\), \(D^{(6)}\sim 10^{-9}\)), yielding \(R(x)<0.10\) (diffusion). $f_G(\theta)$ shows nonlinear drift, varying diffusion (\(\sim 10^{10}\)), and large higher-order terms (\(D^{(4)}\sim 10^{22}\), \(D^{(6)}\sim 10^{33}\)), with \(R(x)\gg 1\) (jump-diffusion).}
    \label{fig:km_gamma_phi_theta}
\end{figure}

The KM coefficients \(D^{(n)}(x)\) for \(n\in\{1,2,4,6\}\) are obtained from the finite-lag-corrected corrected infinitesimal moments via Eq.~\eqref{eq:corrections}.
As a diagnostic, we use the ratio $D^{(4)}/D^{(2)}$ across populated bins to indicate diffusive dynamics that are well described by a Langevin model, whereas large \(R(x)\) signals jump-diffusion.
Fig.~\ref{fig:km_gamma_phi_theta} shows \(D^{(n)}(x)\) and \(R(x)\) for $f_G(\phi)$ (left) and $f_G(\theta)$ (right).
For $f_G(\phi)$, the drift \(D^{(1)}(x)\) is approximately linear and mean-reverting, crossing zero near the equilibrium state; the diffusion \(D^{(2)}(x)\) is stable at \(\sim 4\text{–}7{\times}10^{-4}\) across the state space.
Higher-order coefficients are negligible (\(D^{(4)}\lesssim 4.5{\times} 10^{-7}\), \(D^{(6)}\lesssim 1.6{\times} 10^{-9}\)), and the ratio \(R(x)\) remains below \(0.10\) in all bins, confirming a purely diffusive description.

$f_G(\theta)$ shows a contrasting picture.
The drift \(D^{(1)}(x)\) is nonlinear and state dependent
(\(\approx\!-1.6{\times} 10^{5}\) to \(1.6{\times} 10^{5}\)), and the diffusion varies substantially (\(D^{(2)}(x)\approx 3.0\text{–}9.0\times 10^{10}\)).
The fourth- and sixth-order coefficients attain very large values (\(D^{(4)}\) up to \(\sim 1.2{\times} 10^{22}\); \(D^{(6)}\) up to \(\sim 4.5{\times} 10^{33}\)), producing a diagnostic ratio \(R(x)\) that reaches \(\sim 10^{11}\).
These features unambiguously indicate dominant jump-diffusion dynamics.

The magnitude contrast between $f_G(\phi)$ and $f_G(\theta)$ is substantial: \(D^{(4)}\) differs by roughly
\(10^{29}\) and \(D^{(6)}\) by roughly \(10^{42}\), quantitatively separating the regimes.
A summary for all distribution-parameter pairs appears in Table~\ref{tab:km_summary}; additional coefficient plots for $f_{IG}$, $f_{LN}$, and $f_W$ are provided in Appendix~\ref{app:km_additional}.

We quantify jump activity by inverting the higher-order infinitesimal moments using Eq.~\eqref{eq:jump_params_corrected}.
We also applied the finite time correction in Eq.~\eqref{eq:corrections} to obtain \(F\), the finite time corrected infinitesimal moments.
To obtain global estimates, we apply this procedure to the entire time series (rather than binning in state space), yielding a time-averaged $\hat{\sigma}_\xi$, $\hat{\lambda}$, and the continuous contribution $\hat{D}_{\text{cont}}=F^{(2)}-\hat{\lambda}\hat{\sigma}_\xi^{2}$.
We also report the jump variance share, given by: 
$\hat{f}_{\text{jump}}=\frac{\hat{\lambda}\hat{\sigma}_\xi^{2}}{F^{(2)}}$.

This time-averaged approach maximizes statistical power for the fourth and sixth orders, which are tail-sensitive and noisy.
It provides a single, aggregate characterization of jump activity across the entire trajectory, complementing the spatially resolved KM analysis, which already classifies local dynamics through $D^{(4)}(x)/D^{(2)}(x)$.

Table~\ref{tab:jump_params} lists $\hat{\lambda}$, $\hat{\sigma}_\xi$, and the variance decomposition for all eight series, and Fig.~\ref{fig:variance_decomp} summarizes the jump share.
Results align with the KM classification for Gamma, InvG, and Weib.
The scale parameter $\theta$ exhibits substantial jump variance $\hat{f}_{\text{jump}}\approx 40\%\text{--}63\%$ with jump rates in the range $4\text{--}15$ events per unit time, while the shape parameter $\phi$ shows smaller rates $1\text{--}2$ and modest variance shares $14\%\text{--}22\%$.
These figures confirm that rare, large discontinuities dominate the variability of $\theta$ even when jumps are infrequent, whereas $\phi$ is primarily diffusive.

Log normal departs from this pattern. Although the KM coefficients classify $f_{LN}(\theta)$ as
diffusive, the global inversion produces an elevated $\hat{\lambda}$. This discrepancy reflects the high sensitivity of sixth–order averages to tails and residual corrections under logarithmic parameterization. In such cases, the spatially resolved ratio $D^{(4)}/D^{(2)}$ is the more reliable classifier, and the global inversion should be interpreted cautiously.
The global estimates corroborate the shape-diffusion and scale-jump split observed in the KM analysis and provide practical jump parameters for simulation and risk diagnostics, with the caveat that high-order moment inversions amplify tail effects, especially for $f_{LN}$.

\begin{figure}[t]
    \centering

    \begin{minipage}[t]{0.49\textwidth}
        \centering
        \footnotesize
        \begin{tabular}{lccc}
        \textbf{Distribution} & \textbf{$D^{(2)}$} & \textbf{$D^{(4)}$} & \textbf{Class.} \\
        \hline
        $f_G(\phi)$   & $4$--$7{\times}10^{-4}$      & $\leq 4.5{\times}10^{-7}$   & Diff. \\
        $f_G(\theta)$ & $3$--$9{\times}10^{10}$      & $\leq 1.2{\times}10^{22}$   & Jump \\
        $f_{IG}(\phi)$    & $3.2$--$6.4{\times}10^{-4}$  & $\leq 6.0{\times}10^{-7}$   & Diff. \\
        $f_{IG}(\theta)$  & $0.5$--$1.25{\times}10^{9}$  & $\leq 3.2{\times}10^{18}$   & Jump \\
        $f_{W}(\phi)$    & $2$--$4{\times}10^{-4}$      & $\leq 3.2{\times}10^{-7}$   & Diff. \\
        $f_{W}(\theta)$  & $0.8$--$2.4{\times}10^{10}$  & $\leq 1.2{\times}10^{21}$   & Jump \\
        $f_{LN}(\phi)$    & $1.6$--$2.8{\times}10^{-2}$  & $\leq 1.2{\times}10^{-3}$   & Jump \\
        $f_{LN}(\theta)$  & $1.2$--$1.65{\times}10^{-3}$ & $\leq 4{\times}10^{-6}$     & Diff. \\
        \end{tabular}
        \captionof{table}{Summary of KM coefficient analysis for all
        distribution-parameter combinations.}
        \label{tab:km_summary}
    \end{minipage}
    \hfill
    \begin{minipage}[t]{0.49\textwidth}
        \centering
        \includegraphics[width=\textwidth]{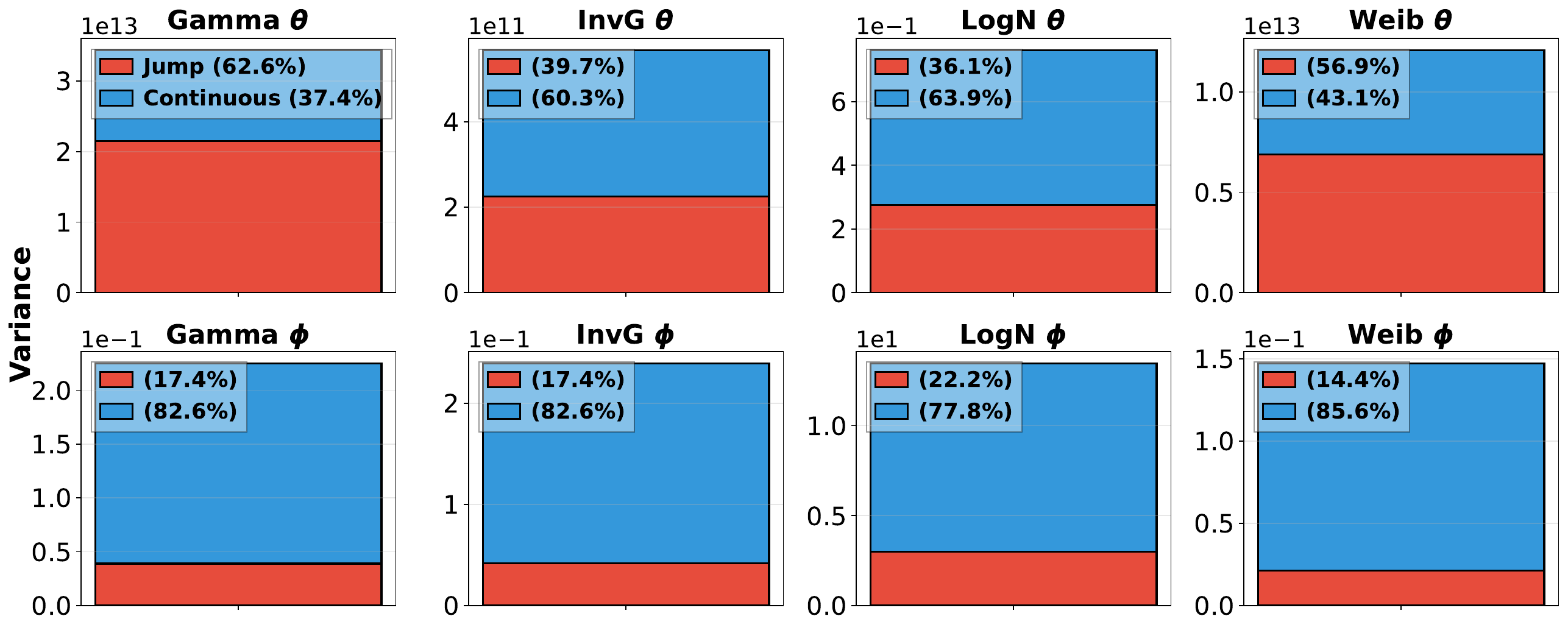}
        \captionof{figure}{Variance decomposition into jump and continuous
        contributions for all distributions. The $\theta$ parameters (top row) show
        significantly higher jump contributions compared to the $\phi$
        parameters (bottom row), consistent with the jump-diffusion
        characterization.}
        \label{fig:variance_decomp}
    \end{minipage}

\end{figure}
\section{Discussion}
\label{sec:discussion}
We begin by discussing the empirical split between the shape and scale dynamics at the intraday level. 
%
For the Gamma, InvG, and Weib distributions, the shape parameter $\phi$ is well described as a diffusion with linear mean reversion, while the scale parameter $\theta$ shows state-dependent jump activity that dominates higher-order statistics.
This split is already visible in Fig.~\ref{fig:cubic_fitting}: the mean intraday profiles of $\theta$ trace a deep U with large amplitude and episodic bursts, whereas $\phi$ varies smoothly with small amplitude.
After detrending, the histograms retain heavier tails for $\theta'$ and near-symmetric cores for $\phi'$.
The KM coefficients confirm this reading.
For $\phi$, the ratio $D^{(4)}/D^{(2)}$ remains below the diffusion threshold across the state space, while for $\theta$ the fourth and sixth orders are large and the diagnostic ratio is far above unity, which is consistent with jump-diffusion dynamics~\cite{Risken1996,tabar2019analysis,RydinGorjao2023}.

LogN shows the opposite qualitative pattern.
In Fig.~\ref{fig:cubic_fitting}, the mean day for LogN has a flatter $\theta$ profile and a stronger slope for $\phi$.
The KM analysis agrees: $f_{LN}(\theta)$ is predominantly diffusive and $f_{LN}(\phi)$ exhibits only weak higher-order activity.
This inversion follows from the logarithmic parameterization.
In LogN, $\phi=\mu$ and $\theta=\sigma$ act in log space, so multiplicative shocks in the original scale become additive in log space.
That regularizes the log-scale $\sigma$ and suppresses spurious jump signatures, while regime changes can still shift the log-mean $\mu$ \cite{tabar2019analysis}.
The mechanism is compatible with a jump-diffusion theory for observable processes~\cite {merton1976jump} and with the KM truncation logic used for classification~\cite{Risken1996,tabar2019analysis}.

Global jump parameters obtained from $M^{(4)}$ and $M^{(6)}$ are useful but sensitive to tails and finite samples. This sensitivity explains the occasional tension between global estimates for $f_{LN}(\theta)$ and its local KM classification. High-order moments amplify rare extremes and can bias inversions for jump rate and amplitude \cite{friedrich2011approaching,anvari2016disentangling}.
In practice, spatially resolved slopes in $\tau$ and the ratio $D^{(4)}/D^{(2)}$ provide the safer classifier because they rely on local small-lag behavior and are less dominated by single outliers \cite{RydinGorjao2023,tabar2019analysis}. Alternative approaches to distinguish jump-diffusion dynamics include parametric L\'evy process models and intermittent process frameworks. While L\'evy processes provide a theoretical foundation for pure-jump behavior with stable distributions, 
and intermittent processes capture regime-switching dynamics, the non-parametric 
KM approach offers flexibility for data that may not conform to predefined 
distribution families. These complementary methodologies are implemented in dedicated software 
libraries~\cite{bhandari2025intlevpy} and illustrated through comprehensive 
empirical studies~\cite{2bzm-t9k1}.

From a market perspective, the split is plausible. The shape parameter, $\phi$, reflects the density and organization of trading flow and adjusts through many small changes.
The scale parameter, $\theta$, tracks liquidity shocks and order-book thinning that arrive discontinuously. The LogN inversion is consistent with a multiplicative structure of monetary flow, where proportional variability appears smooth in log variance and regime shifts move the log mean.
Three elements support the robustness of these findings.
First, de-seasonalizing with a 21-day window and a per-day cubic fit removes deterministic time-of-day structure before inference. Second, explicit verification of the Markov $\tau_M$, combined with finite-lag corrections, controls memory contamination of conditional moments~\cite{gong1992markov,box2015time}.
Third, zone-adaptive binning stabilizes tail estimates and reduces empty bins, which is necessary for fourth- and sixth-order moments~\cite{RydinGorjao2023}.
Remaining limitations include the sample hunger of sixth-order estimates, tail sensitivity to outliers, and residual intraday heterogeneity.
Larger panels, alternative detrending windows, and cross-market replication would strengthen the results.
Online updating and links to order-book features or news timing would sharpen the microstructural interpretation.

\section{Conclusions}
\label{sec:conclusions}

Intraday volume-price parameters are separated into two regimes.
For $f_G$, $f_{IG}$, and $f_W$, the shape parameter follows diffusive dynamics with linear mean reversion.
In contrast, the scale parameter requires a jump-diffusion description supported by large fourth- and sixth-order KM contributions and a high diagnostic ratio $D^{(4)}/D^{(2)}$. $f_{LN}$ is inverted: the log-scale is mostly diffusive, and the log-mean shows only weak higher-order activity, which follows from the logarithmic parameterization that turns multiplicative shocks into additive variation in log space. The KM diagnostic separates these regimes reliably across state bins. At the same time, global jump inversions based on high-order moments should be interpreted with caution, especially for $f_{LN}$, due to tail sensitivity. The observed shape-scale asymmetry is useful in practice. Shape provides a slow structural signal for execution and sizing, while scale should drive volatility controls, stop placement, and stress scenarios. Together, the results describe a diffusive core with a jump-prone shell and a predictable inversion under log normal distribution, aligning empirical intraday behavior with jump-diffusion theory and KM diagnostics.

\backmatter


\section*{Declarations}

\begin{itemize}
    \item \textbf{Conflict of interest/Competing interests:} The authors have no competing interests to declare that are relevant to the content of this article.
    \item \textbf{Data availability:} Data sets used during the current study are available from the corresponding author on reasonable request.
\end{itemize}



\bibliography{bibliography}

@article{Beck2003,
  title = {Superstatistics},
  author = {Beck, C. and Cohen, E. G. D.},
  journal= {Physica A: Statistical Mechanics and its Applications},
  volume = {322},
  pages = {267--275},
  issn = {0378-4371},
  year = {2003},
  doi = {10.1016/S0378-4371(03)00019-0}
}

@Article{Beck2004,
  author = {Beck, C.},
  title = {Superstatistics: theory and applications},
  journal = {Continuum Mechanics and Thermodynamics},
  year = {2004},
  volume = {16},
  number = {3},
  pages = {293--304},
  issn = {1432-0959},
  doi = {10.1007/s00161-003-0145-1},
}

@incollection{sheldon2014Introduction,
title = {Introduction to Probability Models},
editor = {Sheldon Ross},
booktitle = {Introduction to Probability Models (Eleventh Edition)},
publisher = {Academic Press},
edition = {Eleventh Edition},
address = {Boston},
pages = {iii},
year = {2014},
isbn = {978-0-12-407948-9},
doi = {https://doi.org/10.1016/B978-0-12-407948-9.00013-X}
}

@Article{RydinGorjao2021arbitrary,
  author = {Rydin Gorjão, Leonardo and Witthaut, Dirk and Lehnertz, Klaus and Lind, Pedro G.},
  title = {Arbitrary-Order Finite-Time Corrections for the Kramers–Moyal Operator},
  journal = {Entropy},
  volume = {23},
  year = {2021},
  issue = {5},
  pages = {517},
  doi = {10.3390/e23050517}
}

@book{tabar2019analysis,
  author       = {Rahimi Tabar, M. Reza},
  title        = {Analysis and Data-Based Reconstruction of Complex Nonlinear Dynamical Systems},
  subtitle     = {Using the Methods of Stochastic Processes},
  series       = {Understanding Complex Systems},
  publisher    = {Springer Cham},
  year         = {2019},
  doi          = {10.1007/978-3-030-18472-8},
  isbn         = {978-3-030-18471-1},
  isbn10soft   = {978-3-030-18474-2},
  isbn10ebook  = {978-3-030-18472-8},
  pages        = {280},
  edition      = {1},
  address      = {Cham},
  series_issn  = {1860-0832},
  series_eissn = {1860-0840}
}

@article{estevens2017stochastic,
  title={Stochastic modeling of non-stationary financial assets},
  author={Estevens, Joana and Rocha, Paulo and Boto, João P. and Lind, Pedro G.},
  journal={arXiv preprint arXiv:1705.01145},
  year={2017}
}

@article{friedrich2011approaching,
  title={Approaching complexity by stochastic methods: From biological systems to turbulence},
  author={Friedrich, Rudolf and Peinke, Joachim and Sahimi, Muhammad and Rahimi Tabar, M. Reza},
  journal={Physics Reports},
  volume={506},
  pages={87--162},
  year={2011},
  publisher={Elsevier},
  doi={10.1016/j.physrep.2011.05.003}
}

@article{riechers2023discontinuous,
  title={Discontinuous stochastic forcing in Greenland ice core data},
  author={Riechers, Keno and Morr, Andreas and Lehnertz, Klaus and Lind, Pedro G and Boers, Niklas and Witthaut, Dirk and Gorj{\~a}o, Leonardo Rydin},
  journal={arXiv preprint arXiv:2303.06139},
  year={2023}
}

@article{RydinGorjao2023,
  title = {jumpdiff: A Python Library for Statistical Inference of Jump-Diffusion Processes in Observational or Experimental Data Sets},
  author = {Rydin Gorjão, Leonardo and Witthaut, Dirk and Lind, Pedro G.},
  journal = {Journal of Statistical Software},
  volume = {105},
  number = {4},
  pages = {1--22},
  year = {2023},
  doi = {10.18637/jss.v105.i04}
}

@book{Risken1996,
  author    = {Risken, H.},
  title     = {The Fokker-Planck Equation: Methods of Solution and Applications},
  series    = {Springer Series in Synergetics},
  edition   = {2nd},
  publisher = {Springer},
  address   = {Berlin},
  year      = {1996},
  doi       = {10.1007/978-3-642-61544-3}
}

@article{anvari2016disentangling,
  title={Disentangling the stochastic behavior of complex time series},
  author={Anvari, Mehrnaz and Tabar, M Reza Rahimi and Peinke, Joachim and Lehnertz, Klaus},
  journal={Scientific Reports},
  volume={6},
  number={1},
  pages={35435},
  year={2016},
  doi = {10.1038/srep35435},
  publisher={Nature Publishing Group UK London}
}

@article{persio2016jump,
  title={Jump diffusion and $\alpha$-stable techniques for the Markov switching approach to financial time series},
  author={Di Persio, Luca and Jovic, Vukasin},
  journal={arXiv preprint arXiv:1605.05893},
  year={2016}
}

@article{merton1976jump,
  title={Option pricing when underlying stock returns are discontinuous},
  author={Merton, Robert C.},
  journal={Journal of Financial Economics},
  volume={3},
  number={1-2},
  pages={125--144},
  year={1976},
  publisher={Elsevier}
}

@article{black1973pricing,
  title={The pricing of options and corporate liabilities},
  author={Black, Fischer and Scholes, Myron},
  journal={Journal of Political Economy},
  volume={81},
  number={3},
  pages={637--654},
  year={1973},
  publisher={University of Chicago Press}
}

@article{rocha2016evolution,
  title = {Uncovering the evolution of nonstationary stochastic variables: The example of asset volume-price fluctuations},
  author = {Rocha, Paulo and Raischel, Frank and Boto, Jo\~ao P. and Lind, Pedro G.},
  journal = {Phys. Rev. E},
  volume = {93},
  issue = {5},
  pages = {052122},
  numpages = {10},
  year = {2016},
  month = {May},
  publisher = {American Physical Society},
  doi = {10.1103/PhysRevE.93.052122},
  url = {https://link.aps.org/doi/10.1103/PhysRevE.93.052122}
}

@book{gardiner2009stochastic,
  author       = {Gardiner, Crispin},
  title        = {Stochastic Methods},
  subtitle     = {A Handbook for the Natural and Social Sciences},
  series       = {Springer Series in Synergetics},
  publisher    = {Springer Berlin Heidelberg},
  year         = {2009},
  isbn         = {978-3-540-70712-7},
  isbnsoft     = {978-3-642-08962-6},
  edition      = {4},
  pages        = {447},
  series_issn  = {0172-7389},
  series_eissn = {2198-333X},
  address      = {Berlin, Heidelberg}
}

@article{gong1992markov,
  author = {Gong, G. and Tong, H.},
  title = {Testing the Markov property of a time series using conditional entropies},
  journal = {Journal of Time Series Analysis},
  volume = {13},
  number = {6},
  pages = {517--525},
  year = {1992}
}

@article{box2015time,
author = {Wilson, Granville Tunnicliffe},
title = {Time Series Analysis: Forecasting and Control, 5th Edition, by George E. P. Box, Gwilym M. Jenkins, Gregory C. Reinsel, and Greta M. Ljung, 2015. Published by John Wiley and Sons Inc., Hoboken, New Jersey, pp. 712. ISBN: 978-1-118-67502-1},
journal = {Journal of Time Series Analysis},
volume = {37},
number = {5},
pages = {709-711},
doi = {https://doi.org/10.1111/jtsa.12194},
year = {2016}
}

@article{dickey1979distribution,
  author = {Dickey, D.A. and Fuller, W.A.},
  title = {Distribution of the estimators for autoregressive time series with a unit root},
  journal = {Journal of the American Statistical Association},
  volume = {74},
  number = {366},
  pages = {427--431},
  year = {1979}
}

@article{kwiatkowski1992testing,
  author = {Kwiatkowski, D. and Phillips, P.C.B. and Schmidt, P. and Shin, Y.},
  title = {Testing the null hypothesis of stationarity against the alternative of a unit root},
  journal = {Journal of Econometrics},
  volume = {54},
  number = {1--3},
  pages = {159--178},
  year = {1992}
}

@book{klages2008anomalous,
  title        = {Anomalous Transport: Foundations and Applications},
  editor       = {Klages, Rainer and Radons, Günter and Sokolov, Igor M.},
  year         = {2008},
  publisher    = {Wiley-VCH Verlag GmbH \& Co. KGaA},
  isbn         = {9783527407224},
  doi          = {10.1002/9783527622979},
  url          = {https://doi.org/10.1002/9783527622979}
}

@Article{stats6020040,
AUTHOR = {Song, Jiecheng and Ma, Merry},
TITLE = {Climate Change: Linear and Nonlinear Causality Analysis},
JOURNAL = {Stats},
VOLUME = {6},
YEAR = {2023},
NUMBER = {2},
PAGES = {626--642},
URL = {https://www.mdpi.com/2571-905X/6/2/40},
ISSN = {2571-905X},
DOI = {10.3390/stats6020040}
}

@article{10.1063/1.4914547,
    author = {Petelczyc, M. and Żebrowski, J. J. and Orłowska-Baranowska, E.},
    title = {A fixed mass method for the Kramers-Moyal expansion—Application to time series with outliers},
    journal = {Chaos: An Interdisciplinary Journal of Nonlinear Science},
    volume = {25},
    number = {3},
    pages = {033115},
    year = {2015},
    month = {03},
    issn = {1054-1500},
    doi = {10.1063/1.4914547},
    url = {https://doi.org/10.1063/1.4914547},
    eprint = {https://pubs.aip.org/aip/cha/article-pdf/doi/10.1063/1.4914547/14609149/033115\_1\_online.pdf},
}

@article{PhysRevX.14.021037,
  title = {Detection of Approaching Critical Transitions in Natural Systems Driven by Red Noise},
  author = {Morr, Andreas and Boers, Niklas},
  journal = {Phys. Rev. X},
  volume = {14},
  issue = {2},
  pages = {021037},
  numpages = {12},
  year = {2024},
  month = {Jun},
  publisher = {American Physical Society},
  doi = {10.1103/PhysRevX.14.021037},
  url = {https://link.aps.org/doi/10.1103/PhysRevX.14.021037}
}

@article{RENNER_PEINKE_FRIEDRICH_2001, title={Experimental indications for Markov properties of small-scale turbulence}, volume={433},
DOI={10.1017/S0022112001003597},
journal={Journal of Fluid Mechanics},
author={Renner, Christoph and Peinke, Joachim and Friedrich, Rudolf},
year={2001},
pages={383–409}}

@article{madan1990variance,
    author = {Madan, Dilip B. and Carr, Peter P. and Chang, Eric C.},
    title = {The Variance Gamma Process and Option Pricing},
    journal = {European Finance Review},
    volume = {2},
    number = {1},
    pages = {79-105},
    year = {1998},
    month = {04},
    issn = {1382-6662},
    doi = {10.1023/A:1009703431535},
    url = {https://doi.org/10.1023/A:1009703431535}
}

@article{carr2002fine,
 ISSN = {00219398, 15375374},
 URL = {http://www.jstor.org/stable/10.1086/338705},
 author = {Peter Carr and Hélyette Geman and Dilip B. Madan and Marc Yor},
 journal = {The Journal of Business},
 number = {2},
 pages = {305--332},
 publisher = {The University of Chicago Press},
 title = {The Fine Structure of Asset Returns: An Empirical Investigation},
 urldate = {2026-02-17},
 volume = {75},
 year = {2002}
}

@book{Applebaum_2009,
place={Cambridge},
edition={2},
series={Cambridge Studies in Advanced Mathematics},
title={Lévy Processes and Stochastic Calculus},
publisher={Cambridge University Press},
author={Applebaum, David},
year={2009},
collection={Cambridge Studies in Advanced Mathematics}}

@article{https://doi.org/10.1111/1467-9469.00045,
author = {Barndorff-Nielsen, Ole E.},
title = {Normal Inverse Gaussian Distributions and Stochastic Volatility Modelling},
journal = {Scandinavian Journal of Statistics},
volume = {24},
number = {1},
pages = {1-13},
doi = {https://doi.org/10.1111/1467-9469.00045},
url = {https://onlinelibrary.wiley.com/doi/abs/10.1111/1467-9469.00045},
eprint = {https://onlinelibrary.wiley.com/doi/pdf/10.1111/1467-9469.00045},
year = {1997}
}

@article{bhandari2025intlevpy,
  author = {Bhandari, Shailendra and Lencastre, Pedro and Denisov, Sergey and Bystryk, Yurii S. and Lind, Pedro G.},
  title = {IntLevPy: A Python Library to Classify and Model Intermittent and L{\'e}vy Processes},
  journal = {SoftwareX},
  volume = {31},
  pages = {102334},
  year = {2025},
  issn = {2352-7110},
  doi = {10.1016/j.softx.2025.102334},
  url = {https://www.sciencedirect.com/science/article/pii/S2352711025003000}
}

@article{2bzm-t9k1,
  title = {Dynamical law behind eye movements: Distinguishing between L\'evy and intermittent strategies},
  author = {Lencastre, Pedro and Bystryk, Yurii S. and Yazidi, Anis and Denisov, Sergey and Lind, Pedro G.},
  journal = {Phys. Rev. Res.},
  volume = {7},
  issue = {4},
  pages = {043042},
  numpages = {14},
  year = {2025},
  month = {Oct},
  publisher = {American Physical Society},
  doi = {10.1103/2bzm-t9k1},
  url = {https://link.aps.org/doi/10.1103/2bzm-t9k1}
}

\newpage    
\begin{appendices}

\section{Stationarity test }
Stationarity is essential for time-homogeneous KM analysis, ensuring the validity of time-invariant stochastic coefficients \( D^{(n)}(x) \) \cite{10.1063/1.4914547}. We confirm that all detrended fluctuation series exhibit weak stationarity, defined by constant mean, variance, and lag-dependent autocovariance, supporting conditional moment estimation and diffusion classification.

We applied the ADF and in KPSS tests to the detrended fluctuation series for $f_G$, $f_{IG}$, $f_{LN}$, and $f_W$, covering both \( \phi \) and \( \theta \) variables (eight series). The ADF test assesses the presence of a unit root by fitting the regression
\begin{equation}
\Delta x_t = \alpha + \gamma x_{t-1} + \sum_{i=1}^{p} \beta_i \Delta x_{t-i} + \varepsilon_t,
\label{eq:adf}
\end{equation}
rejecting non-stationarity if \( \gamma < 0 \), with p-values less than 0.0001 across all series \cite{dickey1979distribution}. The KPSS test evaluates stationarity around a constant mean using the model
\begin{equation}
x_t = \mu + r_t + \varepsilon_t,
\label{eq:kpss}
\end{equation}
confirming stationarity if residuals remain near zero, with p-values of 0.1 for all series \cite{kwiatkowski1992testing}. These results, summarized in Table~\ref{tab:stationarity}, confirm weak stationarity for all datasets \cite{stats6020040, 10.1063/1.4914547}.

\begin{table}[h]
\centering
\caption{Stationarity test statistics for detrended fluctuation series. All series reject the null hypothesis of a unit root, ADF, and fail to reject the null hypothesis of stationarit,y KPSS, confirming weak stationarity.}
\label{tab:stationarity}
\small
\begin{tabular}{lcccc}
\toprule
\textbf{Distibution} & \textbf{ADF t-stat} & \textbf{ADF p-value} & \textbf{KPSS} & \textbf{KPSS} \\
$f_G(\phi)$ & -13.425 & 0.0000 & 0.010 & 0.1000 \\
$f_G(\theta)$ & -13.389 & 0.0000 & 0.013 & 0.1000 \\
$f_{IG}(\phi)$ & -13.448 & 0.0000 & 0.010 & 0.1000 \\
$f_{IG}(\theta)$ & -14.607 & 0.0000 & 0.011 & 0.1000 \\
$f_{LN}(\phi)$ & -14.542 & 0.0000 & 0.010 & 0.1000 \\
$f_{LN}(\theta)$ & -13.567 & 0.0000 & 0.011 & 0.1000 \\
$f_{W}(\phi)$ & -13.438 & 0.0000 & 0.010 & 0.1000 \\
$f_{W}(\theta)$ & -13.895 & 0.0000 & 0.012 & 0.1000 \\
\end{tabular}
\end{table}


\section{Characterization of detrended fluctuation stochastic dynamics}
\label{app:stochastic_pipeline}

This appendix details the stochastic analysis pipeline used to characterize the dynamics of detrended fluctuation series $\phi'(t)$ and $\theta'(t)$ for eight datasets ($f_G$, $f_{IG}$, $f_{LN}$, and $f_W$ for $\phi$ and $\theta$), as outlined in Section~3. The pipeline encompasses state space discretization, KDE exploration, Markov time analysis, conditional moments estimation, and KM coefficient estimation.

To estimate KM coefficients, the continuous state space is discretized into bins for conditional moment calculations, as the binning strategy significantly impacts moment stability, particularly for higher-order coefficients sensitive to data sparsity and distribution tails. We evaluated four classical histogram-based binning rules: Freedman-Diaconis, Scott, Sturges, and Doane, which estimate bin counts based on sample size, standard deviation, interquartile range, or skewness. These rules assume well-behaved, unimodal distributions, unlike our heavy-tailed, skewed fluctuation series. Table~\ref{tab:bin_counts_comparison} summarizes the bin counts for each dataset.

\begin{table}[h]
\centering
\caption{Estimated number of bins for each dataset using classical binning rules.}
\label{tab:bin_counts_comparison}
\small
\begin{tabular}{lcccc}
\toprule
\textbf{Dataset} & \textbf{Freedman-Diaconis} & \textbf{Scott} & \textbf{Sturges} & \textbf{Doane} \\
$f_G(\phi)$           & 326 & 214 & 16 & 21 \\
$f_G(\theta)$         & 608 & 305 & 16 & 23 \\
$f_{IG}(\phi)$       & 514 & 325 & 16 & 23 \\
$f_{IG}(\theta)$     & 333 & 167 & 16 & 22 \\
$f_{LN}(\phi)$      & 300 & 161 & 16 & 23 \\
$f_{LN}(\theta)$    & 352 & 234 & 16 & 21 \\
$f_{W}(\phi)$         & 295 & 196 & 16 & 21 \\
$f_{W}(\theta)$       & 282 & 144 & 16 & 22 \\
\end{tabular}
\end{table}

\begin{figure}
\centering
\includegraphics[width=0.48\textwidth]{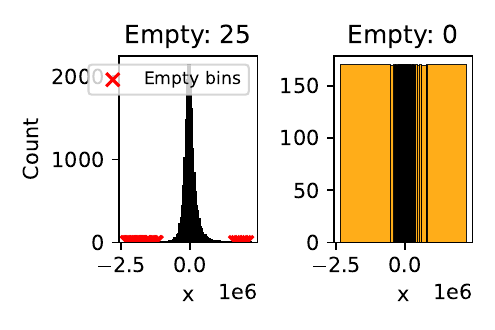}
\caption{Comparison of equal-width binning (left) and quantile-based adaptive binning (right). Equal-width binning results in empty tail bins, while adaptive binning ensures uniform point occupancy.}
\label{fig:equal_width_vs_adaptive}
\end{figure}

Freedman-Diaconis yielded high bin counts (300-608), causing sparse bins (<30 points) in distribution tails, leading to numerical instabilities. Sturges and Doane suggested coarse bins (16-23), missing fine structures needed for KM analysis. Scott’s rule (144-325 bins) was more balanced but still produced 25-60\% empty bins, particularly in tails, due to large fluctuations in financial data. To address this, we implemented a zone-adaptive binning strategy, partitioning the state space into core ($|x| < \sigma$), shoulder ($\sigma \leq |x| < 2\sigma$), and tail ($|x| \geq 2\sigma$) regions, where $\sigma$ is the sample standard deviation. Bins are sized to ensure 500--400 points in core ($n_{\min} = 400$, $n_{\max} = 500$), 300--400 in shoulder ($n_{\min} = 300$, $n_{\max} = 400$), and 200--300 in tail regions ($n_{\min} = 200$, $n_{\max} = 300$). The algorithm processes positive and negative halves separately to handle asymmetry, merging bins below $n_{\min}$ to eliminate empty bins, ensuring stable moment estimates. Figure~\ref{fig:equal_width_vs_adaptive} compares equal-width and adaptive binning.

\begin{figure}
    \centering
    \begin{tabular}{@{}c@{\hspace{0.01\linewidth}}c@{}}
    \fbox{\includegraphics[width=0.46\linewidth]{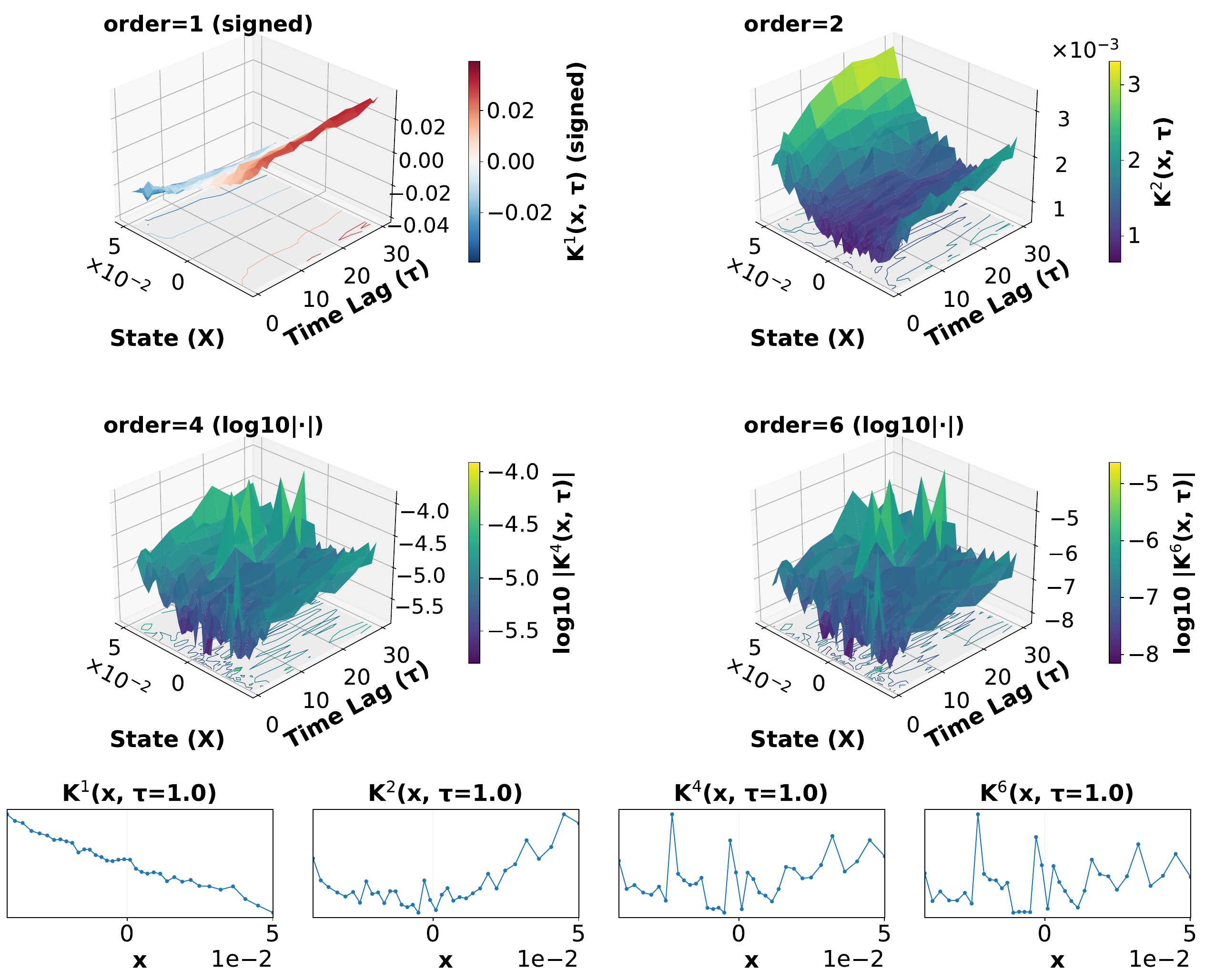}} &
    \fbox{\includegraphics[width=0.46\linewidth]{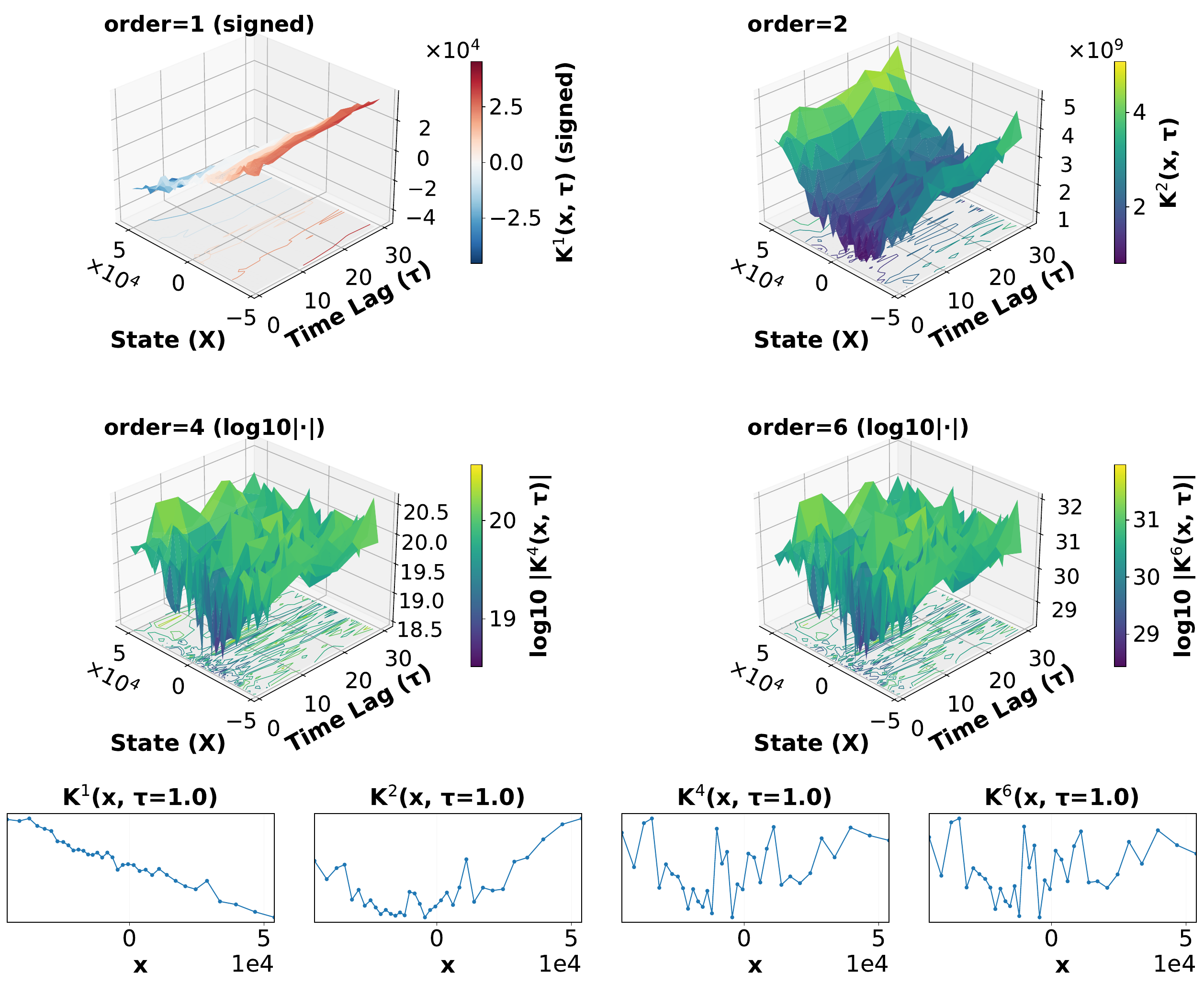}}
    \end{tabular}
    \caption{Raw conditional moments $K^{(n)}(x, \tau)$ for Inverse $f_G(\phi)$ (left) and Inverse $f_G(\theta)$ (right). The layout follows Figure 3 of the main text. $f_{IG}(\phi)$ shows smooth surfaces indicating diffusion. $f_{IG}(\theta)$ shows localized spikes indicating jump-diffusion.}
    \label{fig:invgamma_phi_theta_moments_3d}
\end{figure}

The Markov property, $p(x_{t+\tau} \mid x_t, x_{t-1}, \ldots) = p(x_{t+\tau} \mid x_t)$, is verified to determine the Markov time $\tau_M$, the smallest $\tau$ where the process is approximately memoryless \cite{gong1992markov}. Conditional entropy $H(x_{t+\tau} \mid x_t)$ is computed from joint histograms of $(x_t, x_{t+\tau})$ pairs using \texttt{numpy.histogram2d}:
\begin{equation}
\begin{split}
    H(x_{t+\tau} \mid x_t) = -\sum_{i} p(x_t = i) \sum_{j} p(x_{t+\tau} \\= j \mid x_t = i) \log p(x_{t+\tau} = j \mid x_t = i),
\end{split}
\label{eq:entropy}
\end{equation}
where $p(x_{t+\tau} \mid x_t) = p(x_t, x_{t+\tau}) / p(x_t)$. The Markov time $\tau_M$ is the smallest $\tau$ where the entropy slope $\Delta H = H(\tau+1) - H(\tau)$ falls below 0.005. Complementarily, the PACF measures linear dependence between $x_t$ and $x_{t+\tau}$, controlling for intermediate lags. Using the Yule--Walker method, PACF is computed up to lag 20, with $\tau_M$ identified as the first lag where PACF falls within $\pm 1.96/\sqrt{N}$ ($N \approx 17{,}000$) \cite{box2015time}.

\section{Raw conditional moments for additional distributions}
\label{app:conditional_moments}

This section presents the raw conditional moment analysis $K^{(n)}(x, \tau)$ for $f_{IG}$, $f_{LN}$, and $f_W$ , complementing the Gamma distribution results in Section~\ref{sec:results} of the main text.

\paragraph{Inverse Gamma distribution}

Figure~\ref{fig:invgamma_phi_theta_moments_3d} presents raw conditional moments $K^{(n)}(x, \tau)$ for $f_{IG}(\phi)$ (left) and $f_{IG}(\theta)$ (right). $f_{IG}(\phi)$ shows negligibly higher-order moments with smooth surfaces, indicating pure diffusion. $f_G(\theta)$ displays pronounced localized spikes in fourth and sixth moment surfaces, indicating state-dependent jump-diffusion behavior.

\paragraph{Log Normal distribution}

\begin{figure}
    \centering
    \begin{tabular}{@{}c@{\hspace{0.01\linewidth}}c@{}}
    \fbox{\includegraphics[width=0.46\linewidth]{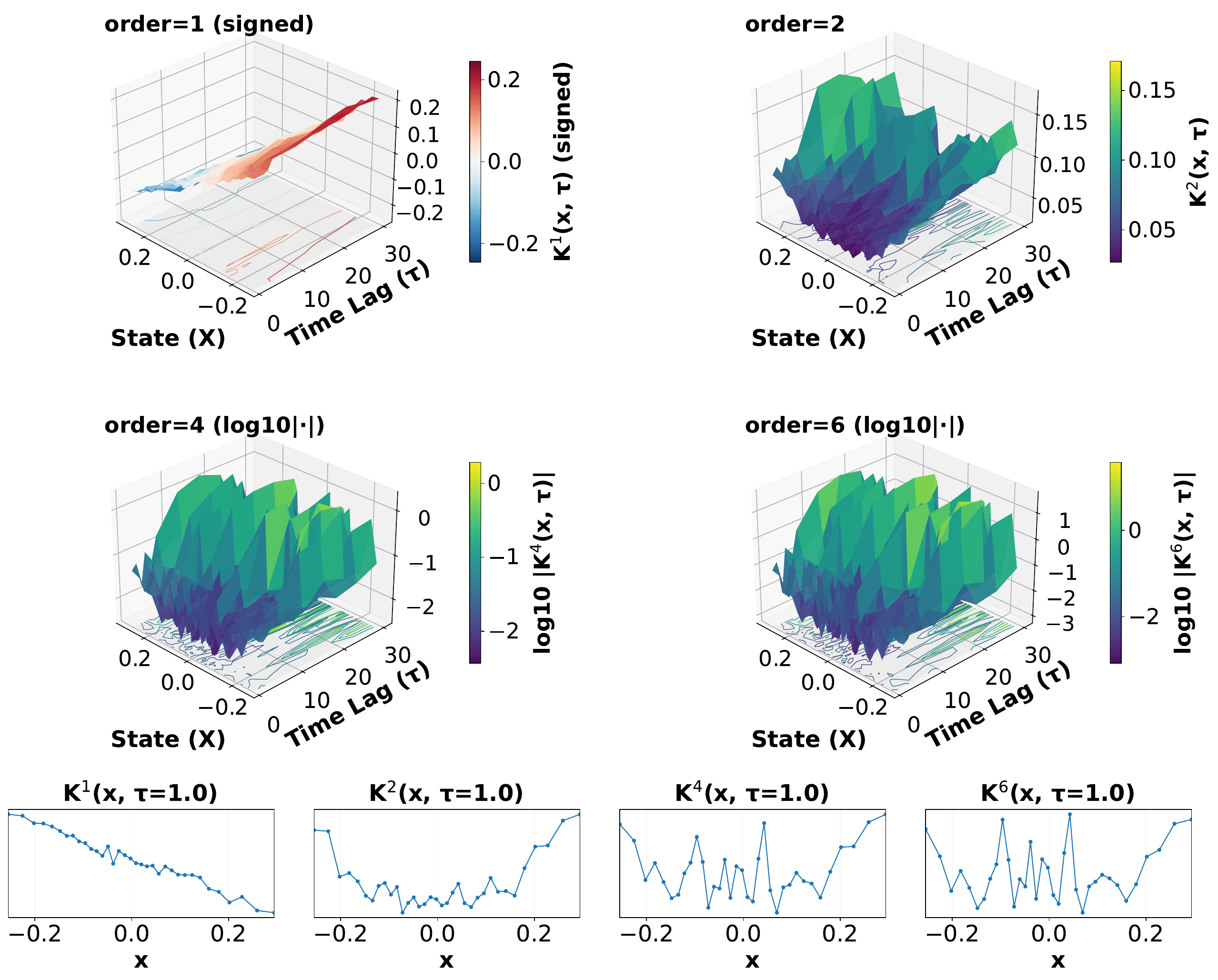}} &
    \fbox{\includegraphics[width=0.45\linewidth]{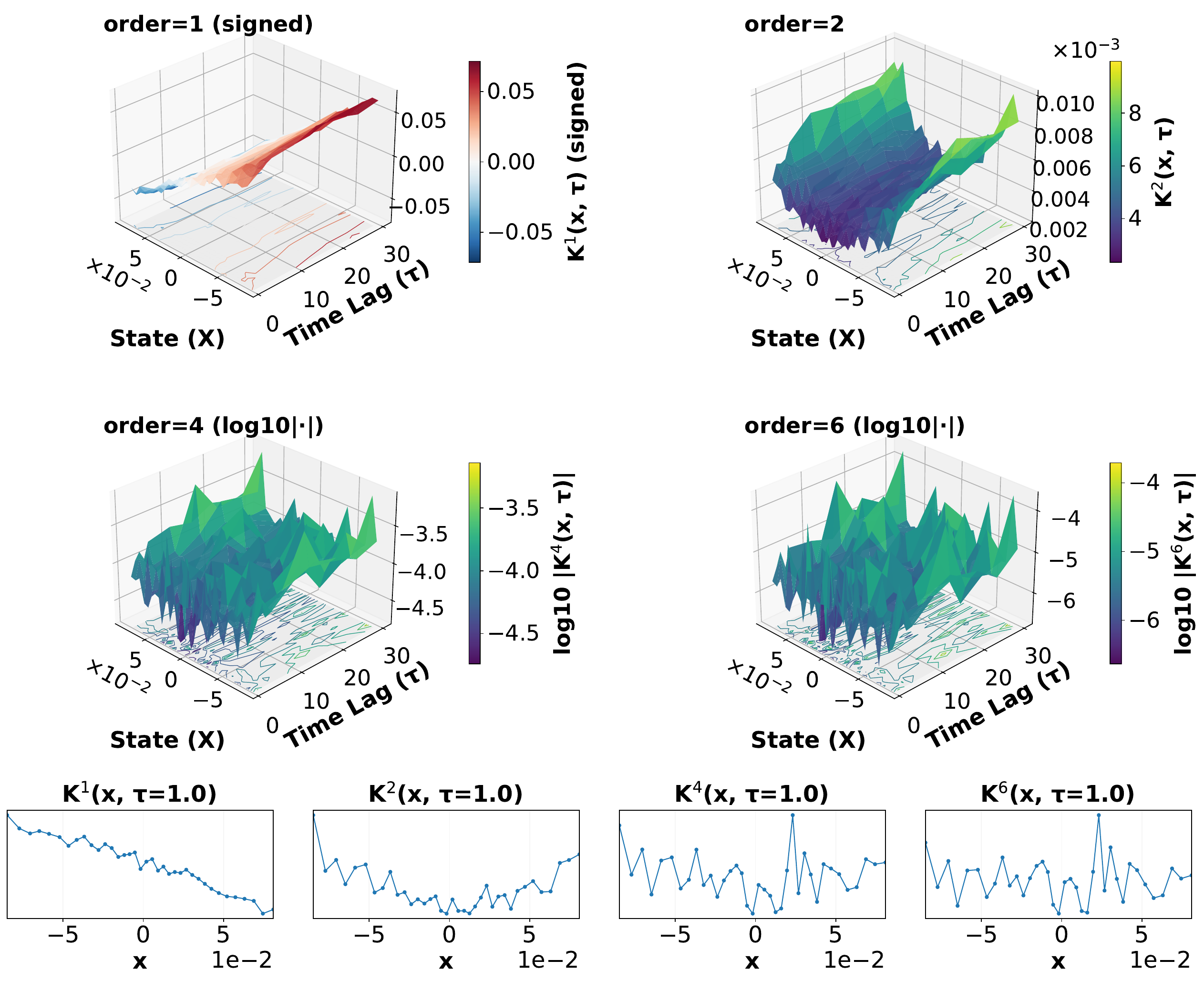}}
    \end{tabular}
    \caption{Raw conditional moments $K^{(n)}(x, \tau)$ for $f_{LN}(\phi)$ (left) and $f_{LN}(\theta)$ (right). The layout follows Figure 3 of the main text. $f_{LN}(\phi)$ shows moderately higher-order moments, suggesting mixed dynamics. $f_{LN}(\theta)$ shows small higher-order moments indicating weak jumps.}
    \label{fig:lognormal_phi_theta_moments_3d}
\end{figure}

Figure~\ref{fig:lognormal_phi_theta_moments_3d} shows raw conditional moments $K^{(n)}(x, \tau)$ for Log-Normal-$\phi$ (left) and $f_{LN}(\theta)$ (right). $f_{LN}(\phi)$ displays intermediate characteristics with moderately higher-order moments. $f_{LN}(\theta)$ shows relatively small higher-order moments compared to other $\theta$ parameters, indicating weak or infrequent jumps.

\paragraph{Weibull distribution}

\begin{figure}
    \centering
    \begin{tabular}{@{}c@{\hspace{0.01\linewidth}}c@{}}
    \fbox{\includegraphics[width=0.46\linewidth]{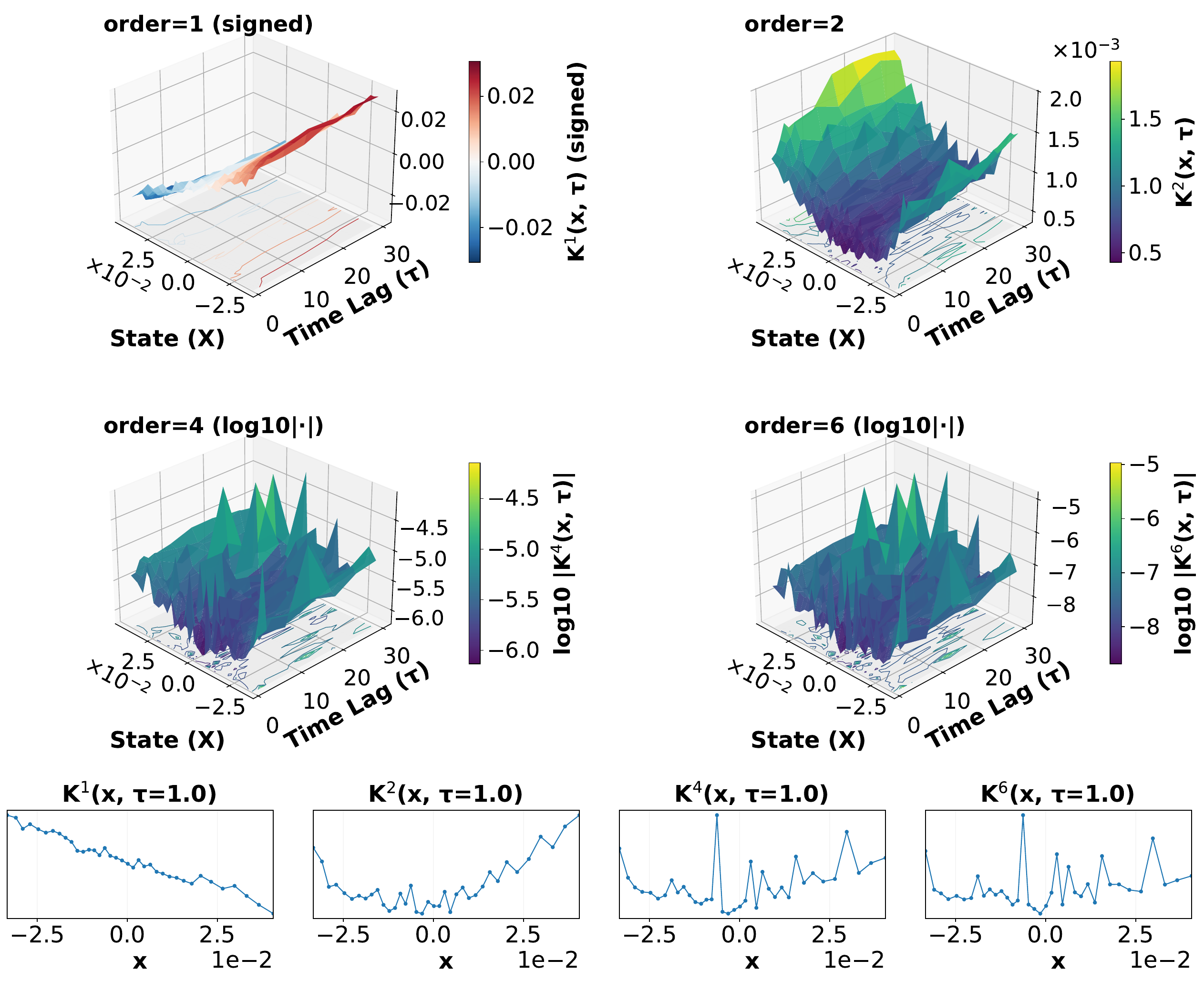}} &
    \fbox{\includegraphics[width=0.46\linewidth]{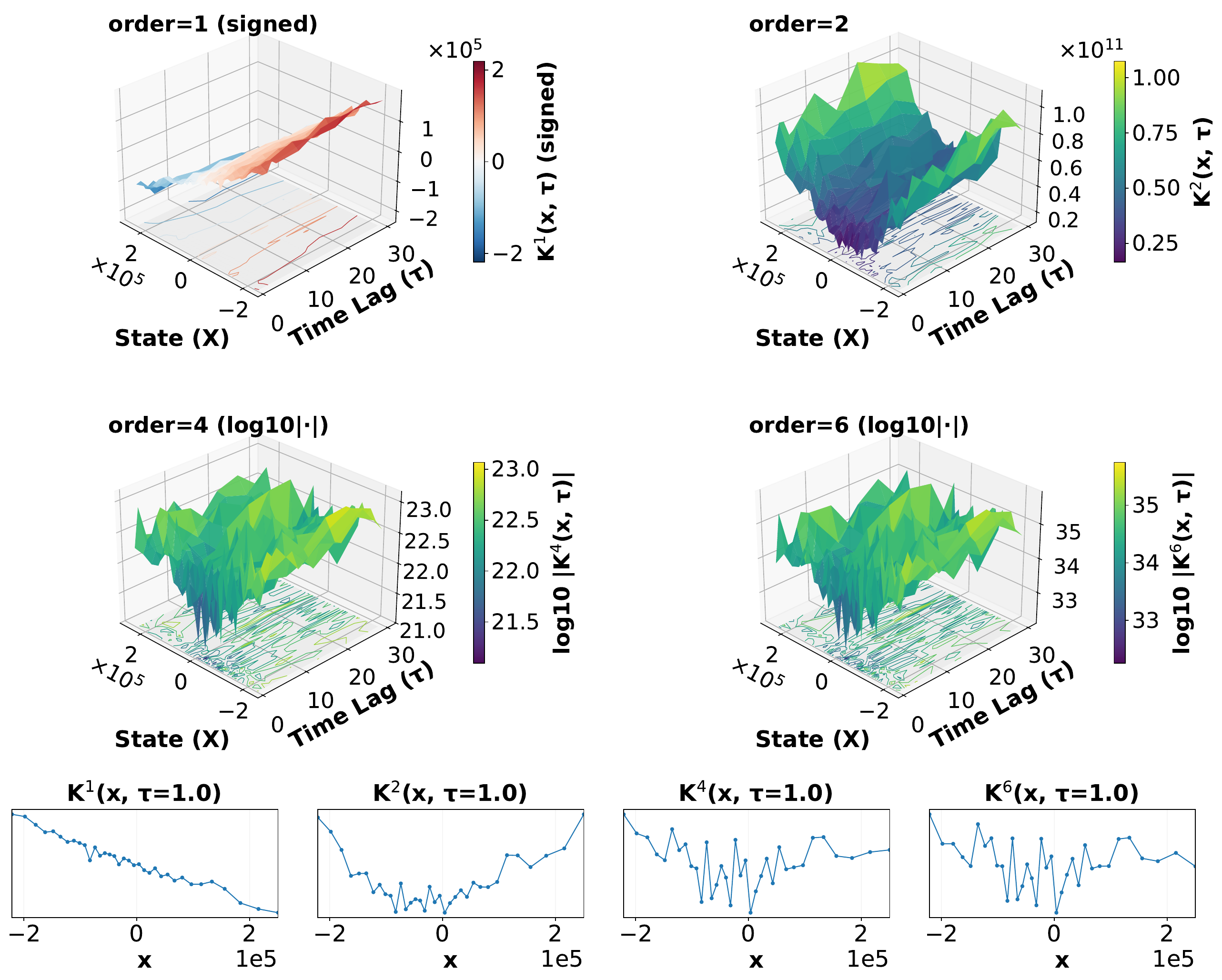}}
    \end{tabular}
    \caption{Raw conditional moments $K^{(n)}(x, \tau)$ for $f_{W}(\phi)$ (left) and $f_{W}(\theta)$ (right). The layout follows Figure 3 of the main text. $f_{W}(\phi)$ shows smooth surfaces indicating pure diffusion. $f_{W}(\theta)$ shows the strongest jump-diffusion signature with extreme moment magnitudes.}
    \label{fig:weibull_phi_theta_moments_3d}
\end{figure}

Figure~\ref{fig:weibull_phi_theta_moments_3d} shows raw conditional moments $K^{(n)}(x, \tau)$ for $f_{W}(\phi)$ (left) and $f_{W}(\theta)$ (right). $f_{W}(\phi)$ exhibits exceptionally smooth surfaces with minimal higher-order moments, representing pure diffusion. $f_{W}(\theta)$ displays the strongest jump-diffusion signature with extreme moment magnitudes and irregular surface topology.

\section{KM coefficients for additional distributions}
\label{app:km_additional}

\begin{table}[t]
\caption{\label{tab:combined_summary}%
Moment magnitudes are reported as $\log_{10}|K^{(n)}(x,\tau{=}1)|$ ranges over populated bins. Classification is based on the
KM diagnostic $R(x)=D^{(4)}(x)/D^{(2)}(x)$: \emph{Diffusive} if median $R<0.1$ with
upper 95\% CI $<0.2$ across bins; otherwise \emph{Jump}.}
\begin{tabular}{lccc}
\textbf{Distribution} &  \textbf{$\log_{10}|K^{(2)}|$} & \textbf{$\log_{10}|K^{(4)}|$} & \textbf{Classification} \\
\hline
$f_G(\phi)$        & $-3$ to $-2$   & $-5.5$ to $-4.0$   & Diffusion \\
$f_G(\theta)$      & $11$ to $12$   & $18.5$ to $20.5$   & Jump \\
$f_{IG}(\phi)$     & $-3$ to $-2$   & $-5.5$ to $-4.0$   & Diffusion \\
$f_{IG}(\theta)$   & $11$ to $12$   & $18.5$ to $20.5$   & Jump \\
$f_{LN}(\phi)$     & $-3$ to $-2$   & $-2$ to $0$        & Weak jump \\
$f_{LN}(\theta)$   & $-3$ to $-2$   & $-4.5$ to $-3.5$   & Diffusion \\
$f_{W}(\phi)$      & $-3$ to $-2$   & $-6.0$ to $-4.5$   & Diffusion \\
$f_{W}(\theta)$    & $11$ to $12$   & $21.0$ to $23.0$   & Jump 
\end{tabular}
\end{table}

This section presents the KM coefficient analysis $D^{(n)}(x)$ for $f_{IG}$, $f_{LN}$, and $f_W$, complementing the Gamma distribution results in the main text. The KM coefficients are estimated from corrected infinitesimal moments $F_n(x)$ via $D^{(n)}(x) = F_n(x)/n!$ as described in the main text.

\paragraph{Inverse Gamma distribution}

\begin{figure}
    \centering
    \begin{tabular}{@{}c@{\hspace{0.01\linewidth}}c@{}}
    \fbox{\includegraphics[width=0.45\linewidth]{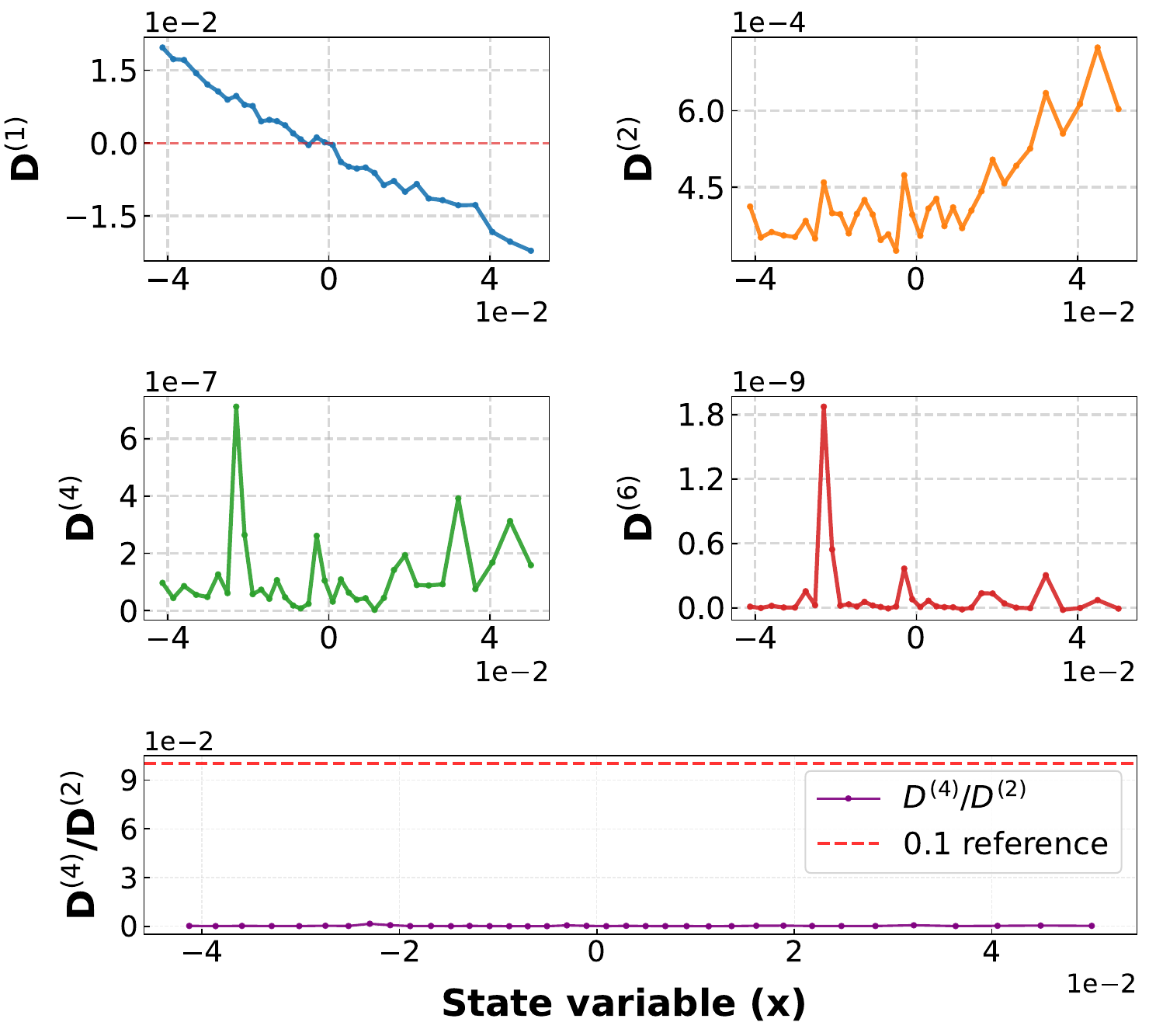}} &
    \fbox{\includegraphics[width=0.45\linewidth]{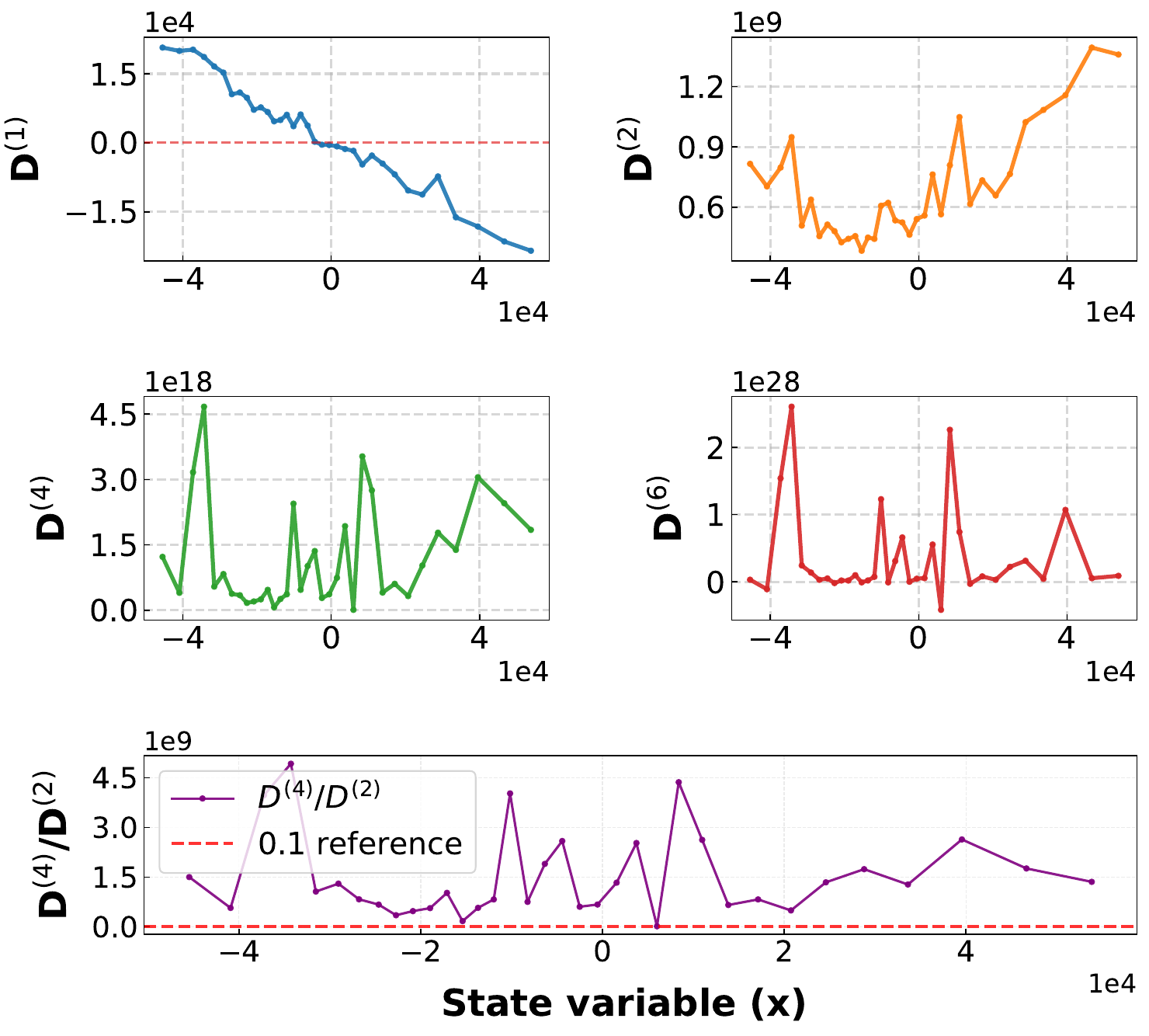}}
    \end{tabular}
    \caption{KM coefficients $D^{(n)}(x)$ for $f_{IG}(\phi)$ (left) and $f_{IG}(\theta)$ (right). The layout follows Figure 4 of the main text. $f_{IG}(\phi)$ shows diffusive behavior with a ratio $< 0.10$. $f_{IG}(\theta)$ exhibits jump-diffusion with ratio $\sim 10^9$.}
    \label{fig:km_invg_phi_theta}
\end{figure}

Figure~\ref{fig:km_invg_phi_theta} shows KM coefficients for $f_{IG}(\phi)$ (left) and $f_{IG}(\theta)$ (right). $f_{IG}(\phi)$ exhibits linear mean-reverting drift, stable diffusion around $\sim 10^{-4}$, and negligible higher-order coefficients ($D^{(4)} \sim 10^{-7}$). The ratio $D^{(4)}/D^{(2)} < 0.10$ confirms diffusion. $f_{IG}(\theta)$ shows nonlinear drift, varying diffusion ($\sim 10^9$), and large higher-order coefficients ($D^{(4)} \sim 10^{18}$). The ratio $\sim 10^9$ indicates jump-diffusion.

\paragraph{Log Normal distribution}

\begin{figure}[htbp]
    \centering
    \begin{tabular}{@{}c@{\hspace{0.01\linewidth}}c@{}}
    \fbox{\includegraphics[width=0.45\linewidth]{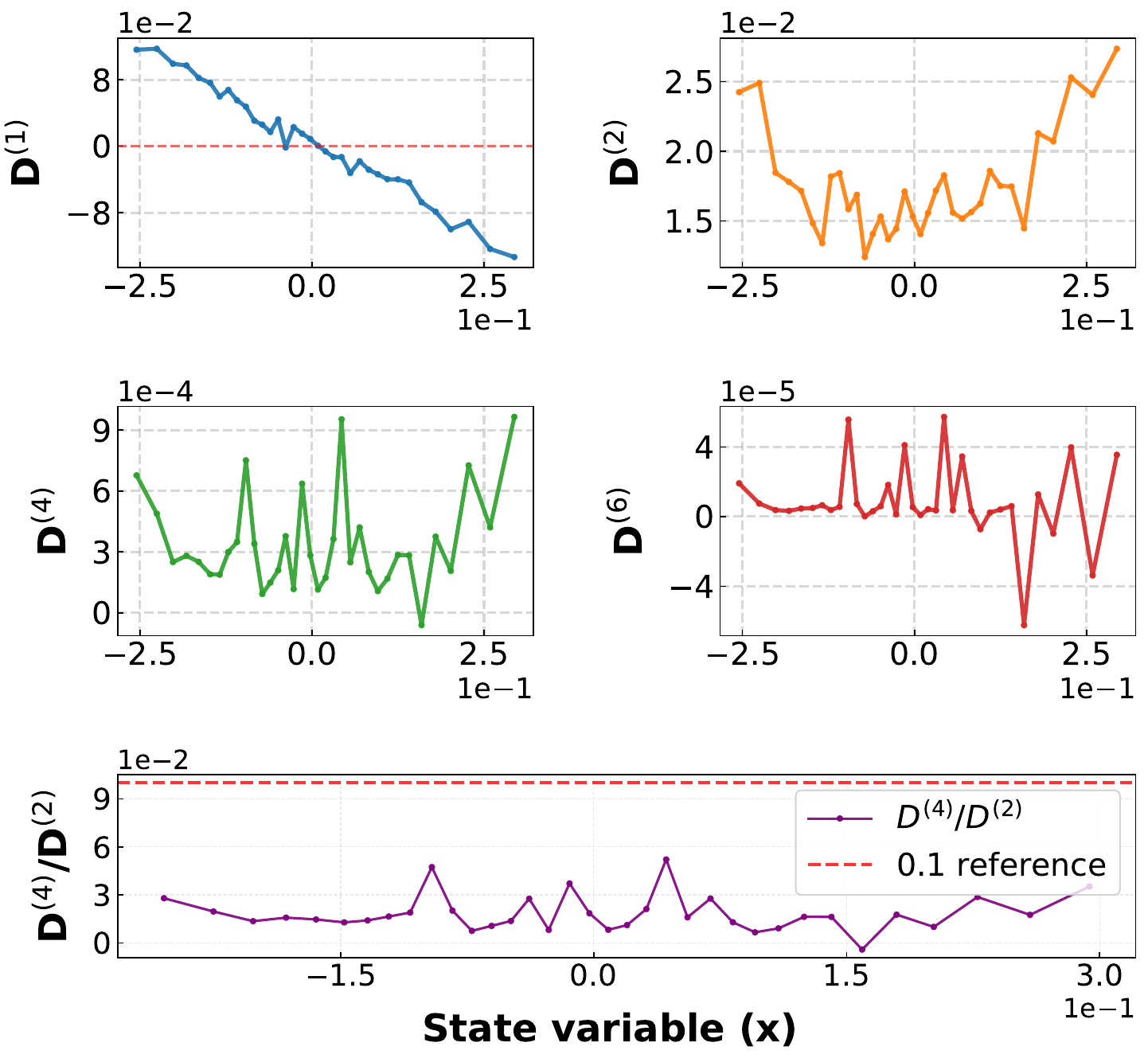}} &
    \fbox{\includegraphics[width=0.45\linewidth]{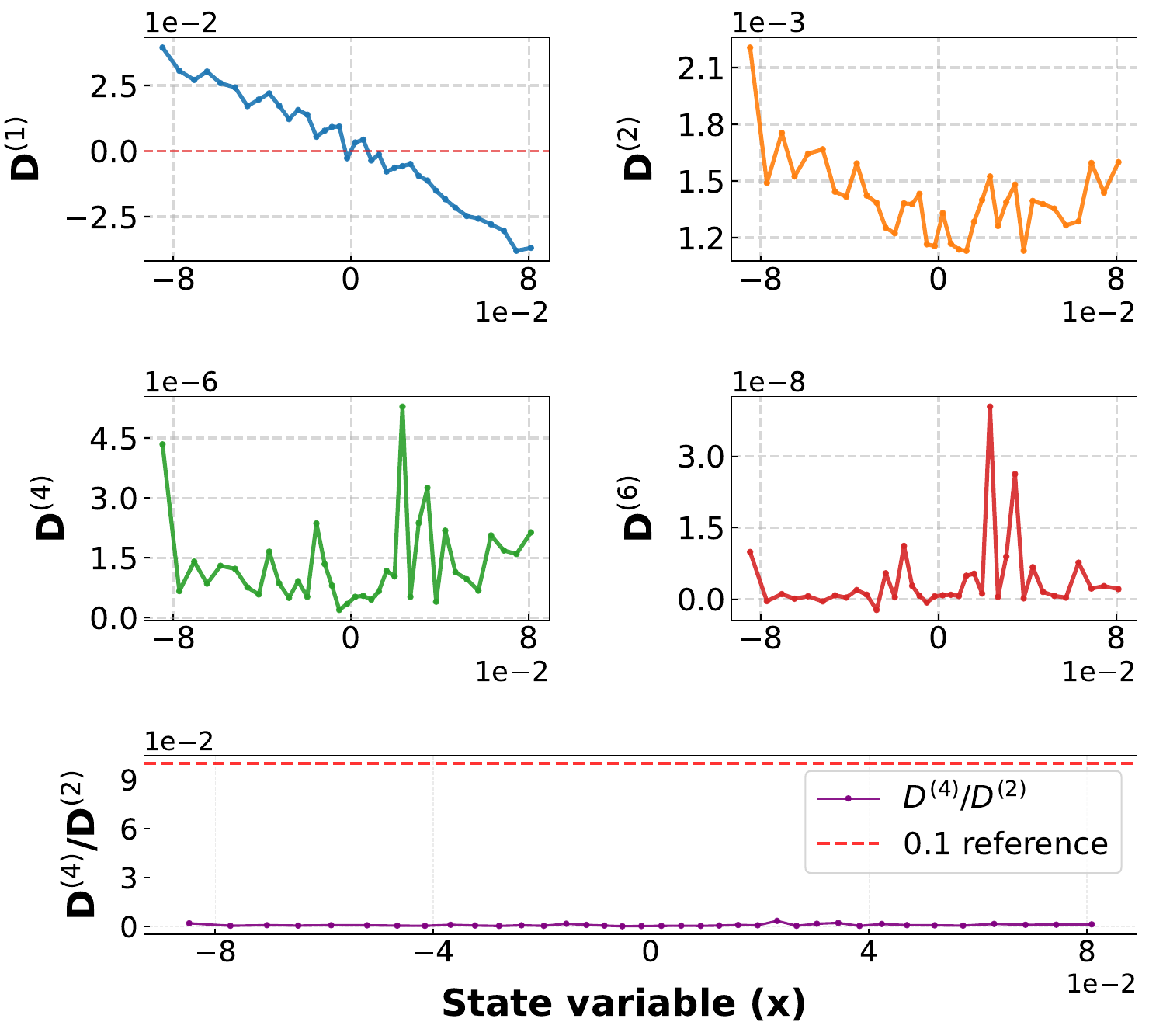}}
    \end{tabular}
    \caption{KM coefficients $D^{(n)}(x)$ for $f_{LN}(\phi)$ (left) and $f_{LN}(\theta)$ (right). The layout follows Figure 4 of the main text. $f_{LN}(\phi)$ shows weak jump-diffusion with an elevated ratio $\approx 0.025$--$0.05$. $f_{LN}(\theta)$ shows diffusion with a ratio $< 0.01$, contrary to other $\theta$ parameters.}
    \label{fig:km_logn_phi_theta}
\end{figure}

Figure~\ref{fig:km_logn_phi_theta} shows KM coefficients for $f_{LN}(\phi)$ (left) and $f_{LN}(\theta)$ (right). $f_{LN}(\phi)$ exhibits linear mean-reverting drift with a larger magnitude than other $\phi$ parameters due to logarithmic scaling. The ratio $D^{(4)}/D^{(2)} \approx 0.025$--$0.05$ is elevated compared to other $\phi$ parameters but remains below 0.10, indicating weak jump-diffusion behavior. $f_{LN}(\theta)$ shows diffusive characteristics with a ratio $< 0.01$, contrary to other $\theta$ parameters. This reversal demonstrates that logarithmic transformation fundamentally alters the stochastic structure.

\paragraph{Weibull distribution}

\begin{figure}[t]
    \centering
    \begin{tabular}{@{}c@{\hspace{0.01\linewidth}}c@{}}
    \fbox{\includegraphics[width=0.45\linewidth]{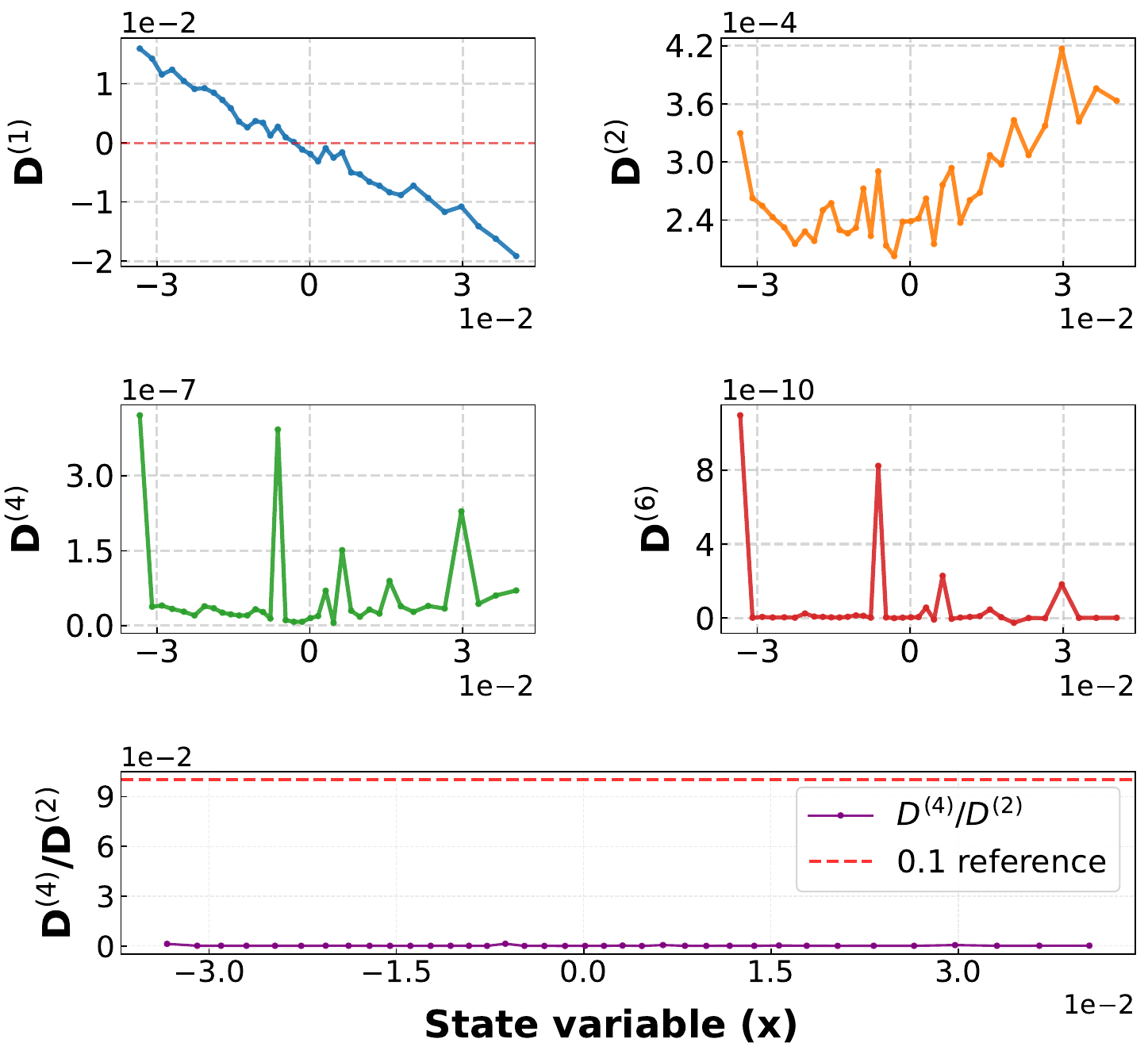}} &
    \fbox{\includegraphics[width=0.45\linewidth]{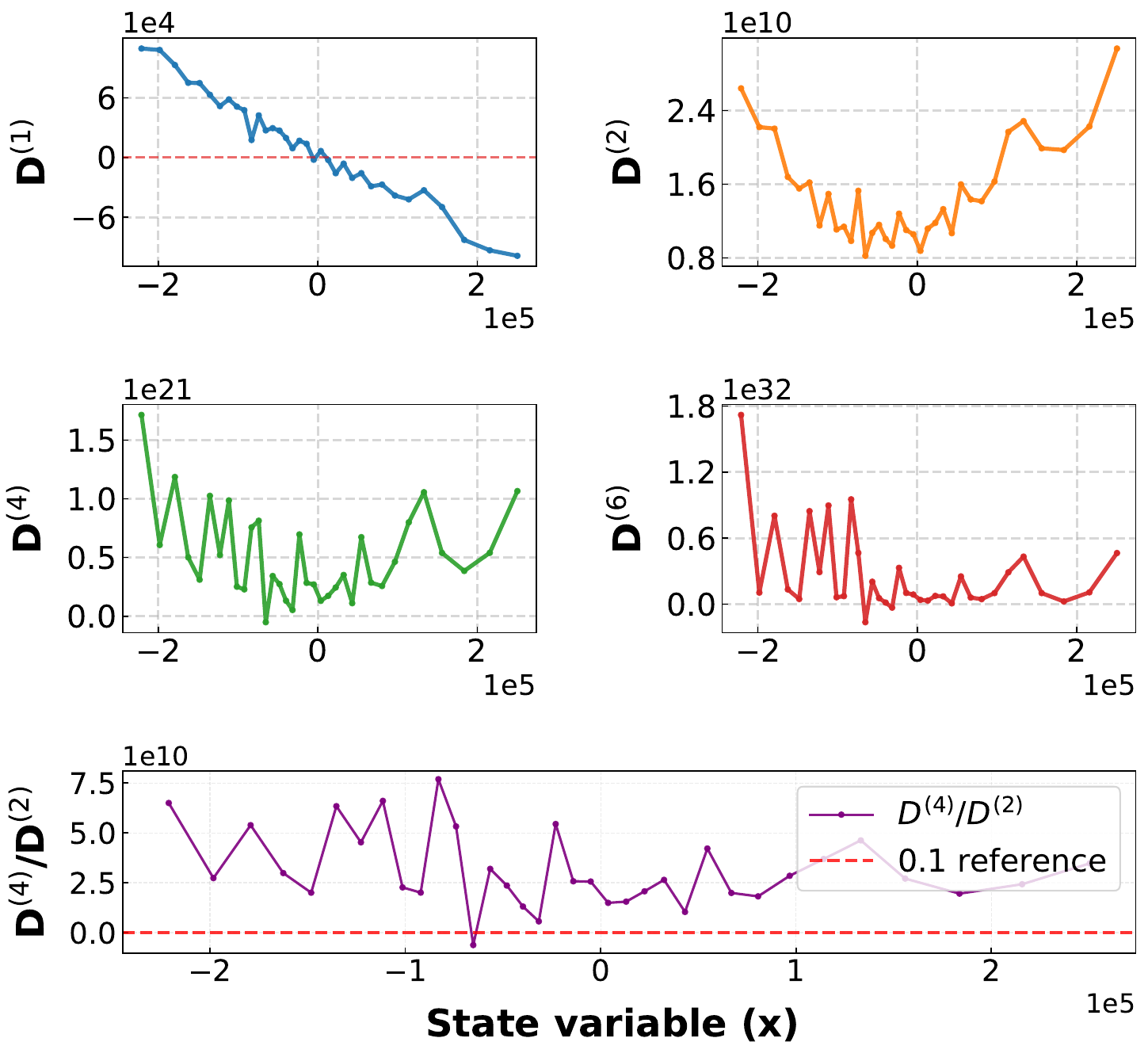}}
    \end{tabular}
    \caption{KM coefficients $D^{(n)}(x)$ for $f_{W}(\phi)$ (left) and $f_{W}(\theta)$ (right). The layout follows Figure 4 of the main text. $f_{W}(\phi)$ shows diffusion with a ratio $< 0.10$. $f_{W}(\theta)$ shows strong jump-diffusion with ratio $\sim 10^{10}$.}
    \label{fig:km_weib_phi_theta}
\end{figure}

Figure~\ref{fig:km_weib_phi_theta} shows KM coefficients for $f_{W}(\phi)$ (left) and $f_{W}(\theta)$ (right). $f_{W}(\phi)$ shows linear mean-reverting drift, stable diffusion around $\sim 10^{-4}$, and negligible higher-order coefficients ($D^{(4)} \sim 10^{-7}$). The ratio $D^{(4)}/D^{(2)} < 0.10$ confirms diffusion. $f_{W}(\theta)$ displays nonlinear drift, varying diffusion ($\sim 10^{10}$), and large higher-order coefficients ($D^{(4)} \sim 10^{21}$). The ratio $\sim 10^{10}$ indicates strong jump-diffusion.

\begin{table}[t]
\caption{\label{tab:jump_params}%
Global jump parameters and variance decomposition for all datasets. 
The jump percentage indicates the fraction of total variance ($M_2$) contributed by jumps.}
\begin{tabular}{lccccc}
Distribution & {${\lambda}$} & ${\sigma_\xi}$ & {${M_2}$} & {${D_{\text{jump}}}$} & {${D_{\text{continuous}}}$} \\
\hline
$f_G(\theta)$ & $14.82$ & $1.21{\times}10^{6}$ & $3.44{\times}10^{13}$ & $2.15{\times}10^{13}$ & $1.29{\times}10^{13}$ \\
$f_{IG}(\theta)$  & $4.04$  & $2.36{\times}10^{5}$ & $5.67{\times}10^{11}$ & $2.25{\times}10^{11}$ & $3.42{\times}10^{11}$ \\
$f_{LN}(\theta)$  & $10.64$ & $1.61{\times}10^{-1}$ & $7.61{\times}10^{-1}$ & $2.74{\times}10^{-1}$ & $4.86{\times}10^{-1}$ \\
$f_W(\theta)$  & $11.87$ & $7.61{\times}10^{5}$ & $1.21{\times}10^{13}$ & $6.87{\times}10^{12}$ & $5.19{\times}10^{12}$ \\
$f_G(\phi)$   & $1.90$  & $1.43{\times}10^{-1}$ & $2.25{\times}10^{-1}$ & $3.90{\times}10^{-2}$ & $1.86{\times}10^{-1}$ \\
$f_{IG}(\phi)$    & $1.89$  & $1.49{\times}10^{-1}$ & $2.39{\times}10^{-1}$ & $4.17{\times}10^{-2}$ & $1.98{\times}10^{-1}$ \\
$f_{LN}(\phi)$    & $0.86$  & $1.86{\times}10^{0}$  & $1.35{\times}10^{1}$  & $2.99{\times}10^{0}$  & $1.05{\times}10^{1}$ \\
$f_W(\phi)$    & $1.16$  & $1.35{\times}10^{-1}$ & $1.47{\times}10^{-1}$ & $2.11{\times}10^{-2}$ & $1.26{\times}10^{-1}$ \\
\end{tabular}
\end{table}

\section{Standard error analysis}
\begin{figure}[t]
    \centering
    \begin{tabular}{@{}c@{\hspace{0.01\linewidth}}c@{}}
    \fbox{\includegraphics[width=0.496\linewidth]{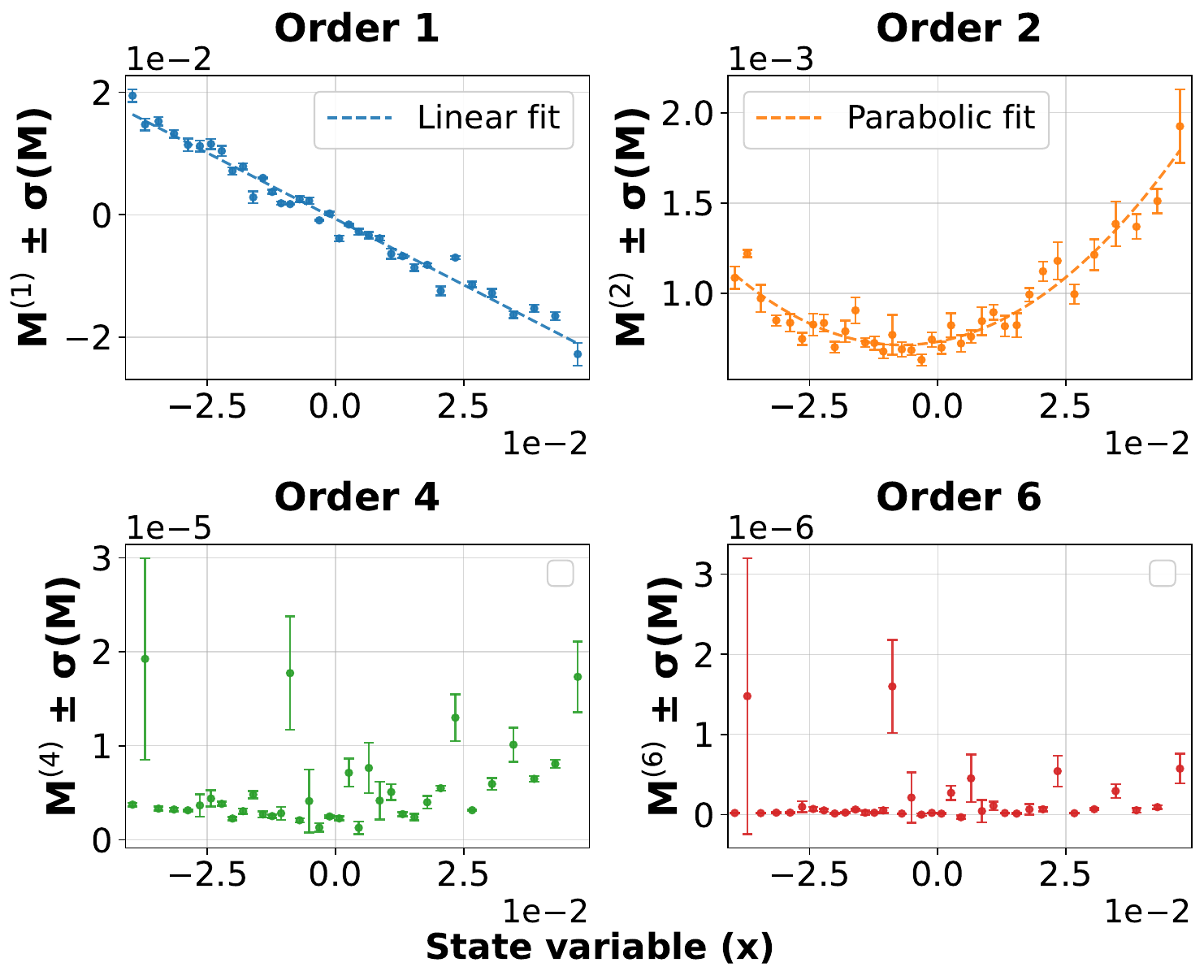}} 
    \fbox{\includegraphics[width=0.496\linewidth]{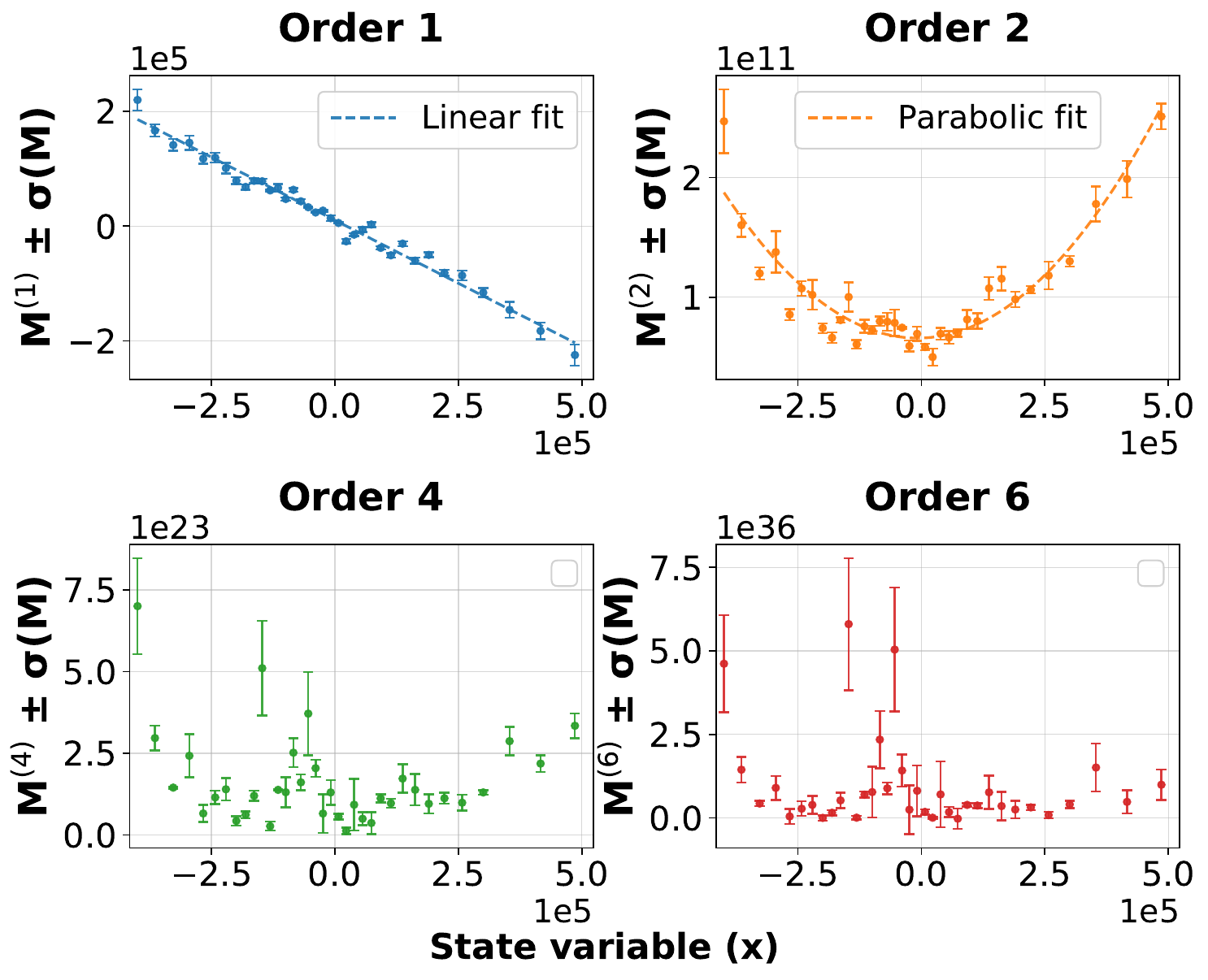}}
    \end{tabular}
    \caption{Standard errors of $M^{(n)}(x)$ for $f_G(\phi)$ (top) and $f_G(\theta)$ (bottom). Error bars show $M^{(n)}(x)\pm\sigma_{\mathrm{w}}$. Errors are smaller in central bins and larger in the tails, and they increase with order $n$. The diffusion versus jump-diffusion classification is unchanged.}
    \label{fig:error_amplitudes}
\end{figure}

\begin{figure}[h]
    \centering
    \begin{tabular}{@{}c@{\hspace{0.01\linewidth}}c@{}}
    \fbox{\includegraphics[width=0.496\linewidth]{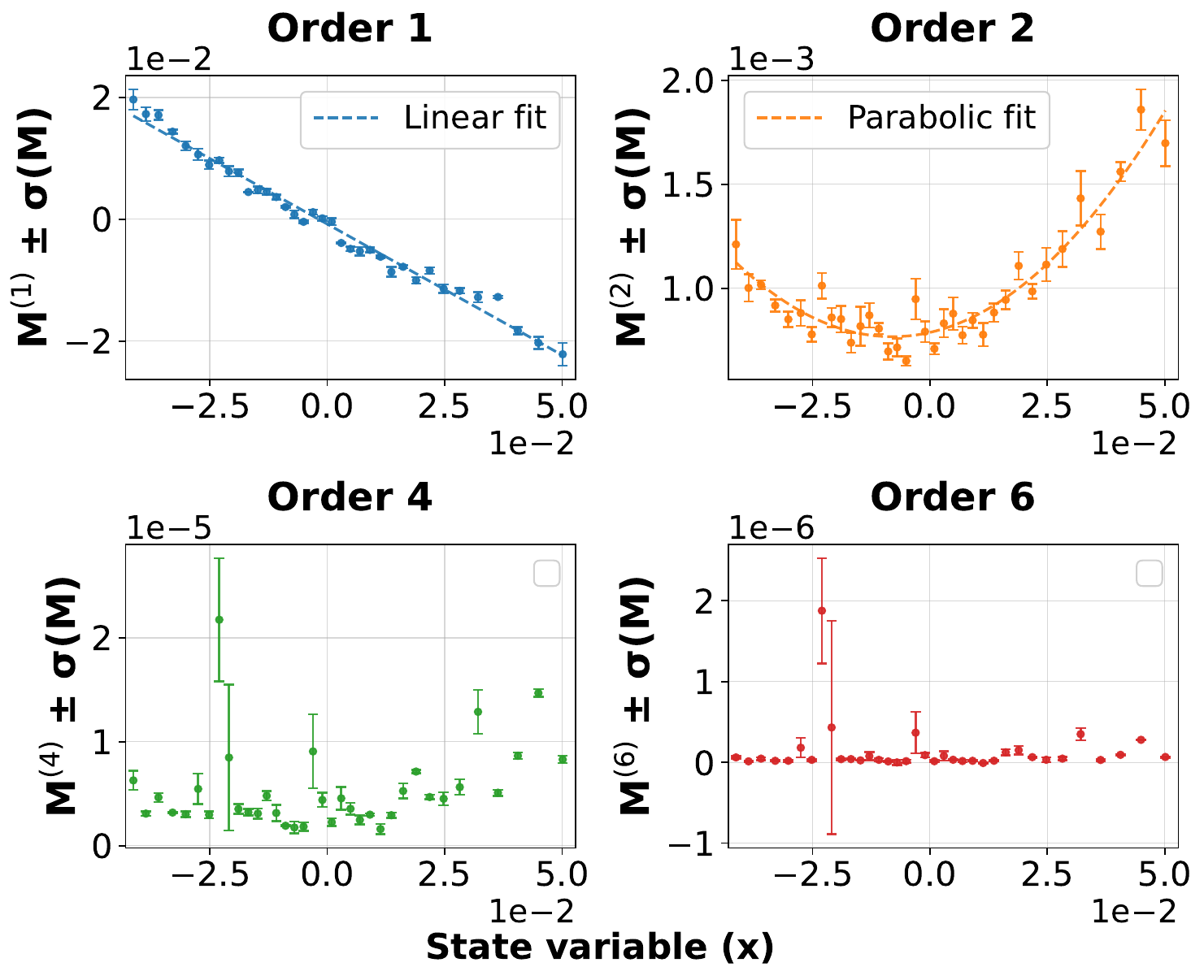}} 
    \fbox{\includegraphics[width=0.496\linewidth]{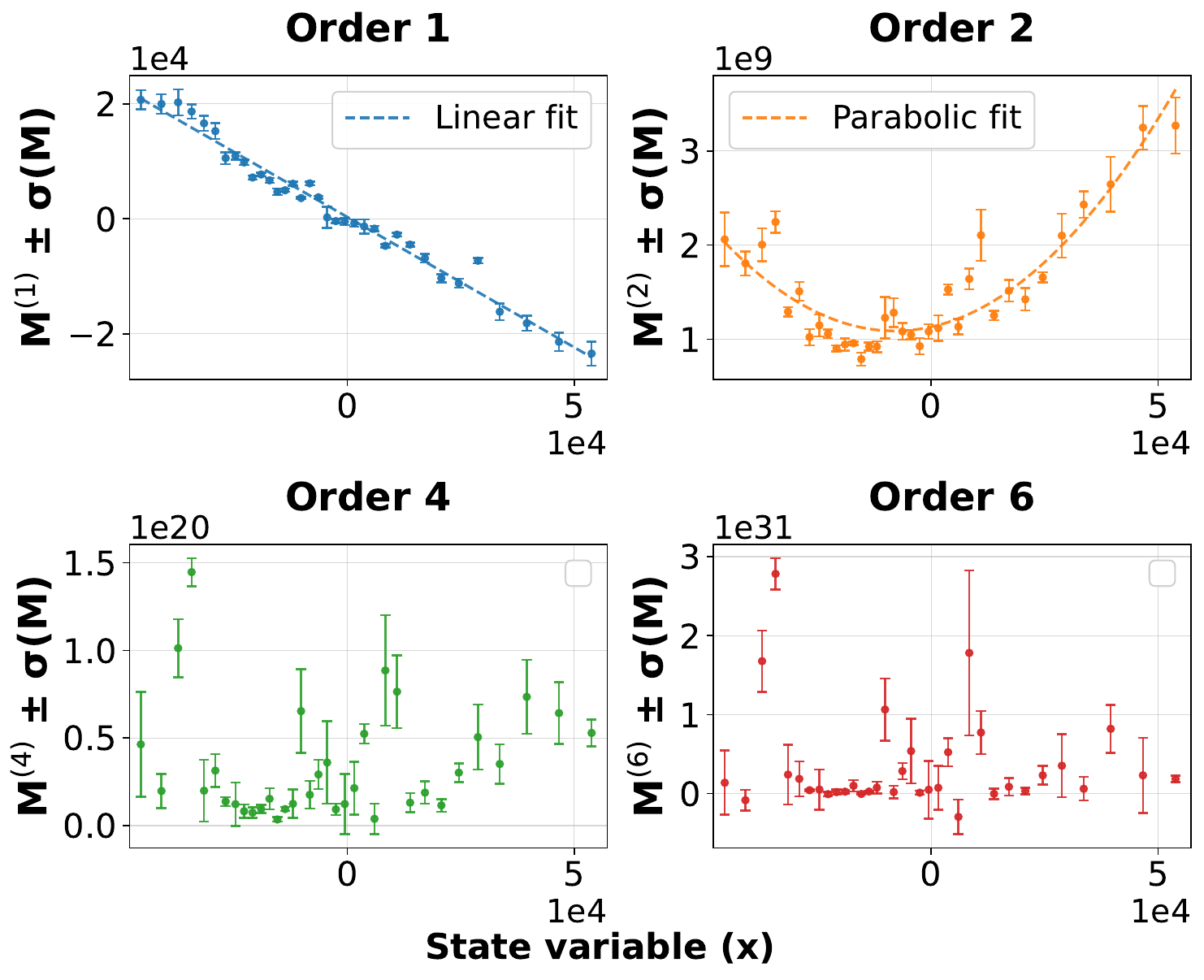}}
    \end{tabular}
    \caption{Standard errors of $M^{(n)}(x)$ for $f_{IG}(\phi)$ (top) and $f_{IG}(\theta)$ (bottom). Error bars show $M^{(n)}(x)\pm\sigma_{\mathrm{w}}$. Errors are smaller in central bins and larger in the tails, and they increase with order $n$. The diffusion versus jump-diffusion classification is unchanged.}
    \label{fig:error_amplitudes1}
\end{figure}

Figure~\ref{fig:error_amplitudes} and \ref{fig:error_amplitudes1} presents standard errors of the fitted infinitesimal moments $\sigma\!\left(M^{(n)}\right)$, across the state bins.
Errors are smaller in central bins due to higher sample occupancy and larger in tail regions due to data sparsity, reflecting the heteroskedastic error structure.
Higher-order moments show larger relative uncertainties due to sensitivity to rare events.
To obtain fit quality, we use a quality-weighted error that downweights poorly explained linear fits of $K^{(n)}(x,\tau)/(\tau\Delta t)$ versus $\tau\Delta t$, given by
\begin{equation}
    \sigma_{\text{w}}(M^{(n)}) \;=\; \sigma_{\text{intercept}}\sqrt{1-R^{2}},
\end{equation}
where $R^{2}$ is the coefficient of determination of the local regression and $\sigma_{\text{intercept}}$ is the standard error of the intercept.
Despite these uncertainties, the regime classification is stable.
For all $\phi$ series we find $D^{(4)}/D^{(2)} \approx 10^{-3}$–$10^{-2}$ across populated bins, well below the diffusion threshold.
For all $\theta$ series the ratio attains $10^{9}$--$10^{11}$, far above unity. The contrast spans many orders of magnitude and is unaffected by the error levels reported here.

Higher-order KM estimates are sensitive to rare tail events and can depend on preprocessing. To test robustness, we re-ran the complete KM pipeline using three moving-average detrending windows: 10 trading days (short), 21 trading days (baseline used in the manuscript), and 42 trading days (long). For each window we repeated the identical steps: adaptive binning, estimation of raw conditional moments $K^{(n)}(x,\tau)$ for $\tau\in\{1,\dots,6\}$, regression-based extraction of infinitesimal moments with the same finite-lag correction, and KM coefficient estimation $D^{(n)}(x)=F^{(n)}(x)/n!$. We then computed the diagnostic ratio $R(x)=D^{(4)}(x)/D^{(2)}(x)$ and averaged it over populated bins (see Table \ref{tab:window_sensitivity_ratio}).
\begin{table}[t]
\centering
\caption{Sensitivity of the KM diagnostic ratio $D^{(4)}/D^{(2)}$ to detrending window length. Values are means across populated state bins. The diffusion versus jump--diffusion classification is unchanged across window choices.}
\label{tab:window_sensitivity_ratio}
\small

\begin{tabular}{lcccc}
\toprule
\textbf{Series} & \textbf{10-day} & \textbf{21-day} & \textbf{42-day} & \textbf{Classification} \\
\midrule
$f_G(\phi)$      & $4.61{\times}10^{-4}$ & $4.78{\times}10^{-4}$ & $4.06{\times}10^{-4}$ & Diffusion \\
$f_G(\theta)$    & $1.46{\times}10^{11}$ & $1.11{\times}10^{11}$ & $1.06{\times}10^{11}$ & Jump--Diffusion \\
$f_{IG}(\phi)$   & $4.51{\times}10^{-4}$ & $4.48{\times}10^{-4}$ & $3.25{\times}10^{-4}$ & Diffusion \\
$f_{IG}(\theta)$ & $2.13{\times}10^{9}$  & $2.22{\times}10^{9}$  & $1.77{\times}10^{9}$  & Jump--Diffusion \\
$f_{LN}(\phi)$   & $6.04{\times}10^{-2}$ & $6.02{\times}10^{-2}$ & $3.72{\times}10^{-2}$ & Diffusion \\
$f_{LN}(\theta)$ & $1.24{\times}10^{-3}$ & $1.22{\times}10^{-3}$ & $1.05{\times}10^{-3}$ & Diffusion \\
$f_W(\phi)$      & $3.48{\times}10^{-4}$ & $3.26{\times}10^{-4}$ & $2.88{\times}10^{-4}$ & Diffusion \\
$f_W(\theta)$    & $3.76{\times}10^{10}$ & $3.79{\times}10^{10}$ & $3.36{\times}10^{10}$ & Jump--Diffusion \\
\bottomrule
\end{tabular}
\end{table}
Across a four-fold change in detrending horizon (10 to 42 days), the magnitude separation between diffusive series ($R\sim 10^{-4}$--$10^{-2}$) and jump--diffusive series ($R\sim 10^{9}$--$10^{11}$) remains many orders of magnitude. Therefore, the diffusion-versus-jump-diffusion classification is robust to the choice of detrending window.


\end{appendices}

\end{document}